\documentclass[10pt,twocolumn,letterpaper]{article}

\usepackage{wacv}              

\usepackage[most]{tcolorbox}
\usepackage{xcolor}
\usepackage[most]{tcolorbox}
\tcbuselibrary{skins}
\usepackage{xcolor}
\usepackage{graphicx}   
\newcommand{\jicon}[1]{\raisebox{-0.35ex}{\includegraphics[height=2.2ex]{logos/#1.pdf}}}

\definecolor{promptfill}{HTML}{F4F6FB}   
\definecolor{promptaccent}{HTML}{3A4CA8} 
\definecolor{promptrule}{HTML}{C7D0E8}   
\definecolor{promptvar}{HTML}{C0392B}    
\usepackage{makecell}
\usepackage[most]{tcolorbox}
\tcbuselibrary{skins}
\usepackage{xcolor}
\usepackage{caption}
\definecolor{promptfill}{HTML}{EAF0FB}
\definecolor{promptrule}{HTML}{B7C6E6}
\definecolor{promptvar}{HTML}{2E3FB0}

\newtcolorbox{promptbox}[1]{
  enhanced,
  colback=promptfill,
  colframe=promptrule,
  boxrule=0.8pt,
  arc=6pt, outer arc=6pt,
  left=12pt, right=12pt, top=18pt, bottom=12pt,
  fontupper=\small,
  overlay={
    \node[anchor=west,
          fill=white, draw=black!70, line width=0.8pt,
          rounded corners=4pt,
          inner xsep=8pt, inner ysep=3pt,
          font=\bfseries\small]
      at ([xshift=14pt, yshift=0pt]frame.north west) {#1};
  },
}
\newcommand{\pv}[1]{\textcolor{promptvar}{\textbf{\{#1\}}}}
\usepackage{lipsum}

\definecolor{exfill}{HTML}{F7F8FA}
\definecolor{exrule}{HTML}{C8D0DE}
\definecolor{exaccent}{HTML}{2E3FB0}
\definecolor{exhumanbg}{HTML}{EEF3EC}
\definecolor{exvlmbg}{HTML}{EDF1FB}

\newtcolorbox{examplebox}[1]{
  enhanced,
  colback=exfill, colframe=exrule, boxrule=0.8pt,
  arc=5pt, outer arc=5pt,
  left=10pt, right=10pt, top=10pt, bottom=8pt,
  fonttitle=\bfseries\small, coltitle=white,
  colbacktitle=exaccent,
  title={#1},
  attach boxed title to top left={xshift=10pt, yshift=-\tcboxedtitleheight/2},
  boxed title style={colback=exaccent, colframe=exaccent, arc=3pt},
}
\newcommand{\exlabel}[2]{\smallskip\noindent\colorbox{#2}{\small\bfseries\,#1\,}\par\smallskip}
\newcommand{\hcrit}[1]{\textit{``#1''}}

\definecolor{wacvblue}{rgb}{0.21,0.49,0.74}
\usepackage[pagebackref,breaklinks,colorlinks,allcolors=wacvblue]{hyperref}

\def\wacvPaperID{1923} 
\def\confName{WACV}
\def\confYear{2027}

\title{Can MLLMs Critique Like Humans? Evaluating Open-Ended Aesthetic Reasoning in Multimodal Large Language Models}

\author{Sajjad Ghiasvand\thanks{UCSB, \texttt{\small \{sajjad,\,alizadeh,\,ramtin\}@ucsb.edu}}
\and
Maryam Amirizaniani\thanks{University of Washington, {\tt\small amaryam@uw.edu}}
\and
Haniyeh Ehsani Oskouie\thanks{UCLA, {\tt\small haniyeh@cs.ucla.edu}}
\and
Mahnoosh Alizadeh\footnotemark[1]
\and
Ramtin Pedarsani\footnotemark[1]
}
\begin{document}
\maketitle
\begin{abstract}

Open-ended aesthetic critique is a challenge for multimodal large language
models (MLLMs): it has no single correct answer, and most aesthetic evaluation
measures models against numeric scores rather than the written critiques
people actually give. We ask whether MLLM critiques are close to human ones, scoring eight
open-weight MLLMs from $7$B to $397$B, plus GPT-5.5, against multiple ranked
human critiques for each of $1{,}227$ \texttt{r/photocritique} posts under
eight prompt conditions. Reference-based similarity gives a misleading picture. In absolute terms the stricter lexical and
learned metrics align only weakly with human critiques while a coarse
embedding cosine reports broad topical overlap, yet requesting shorter
critiques raises those scores and withholding the image barely changes them:
the similarity reflects length, the post text, and a stable critiquing style
more than image-specific observation. An LLM judge sharpens the question rather than settling it: in the primary
condition all four judges prefer the frontier models' critiques to the human
ones, but on the $7$--$8$B models they diverge wildly, from $9\%$ to $81\%$
preference on identical pairs. Asked instead how similar each pair is in
substance, those judges and two human annotators agree, rating every model
between $1.81$ and $2.59$ on a $1$--$5$ scale, close to ``mostly
different''. Behaviorally, the models diverge in ways
the scores do not surface: they cover nearly every aesthetic aspect where
humans are selective and repeat themselves across critiques of one photo,
even when prompted to write at human length. We argue that reference-based similarity rewards a fluent, comprehensive
critique style rather than the selectivity and specificity of human critique.
\end{abstract}

\section{Introduction}

\begin{figure*}[t]
\centering
\includegraphics[width=0.7\textwidth]{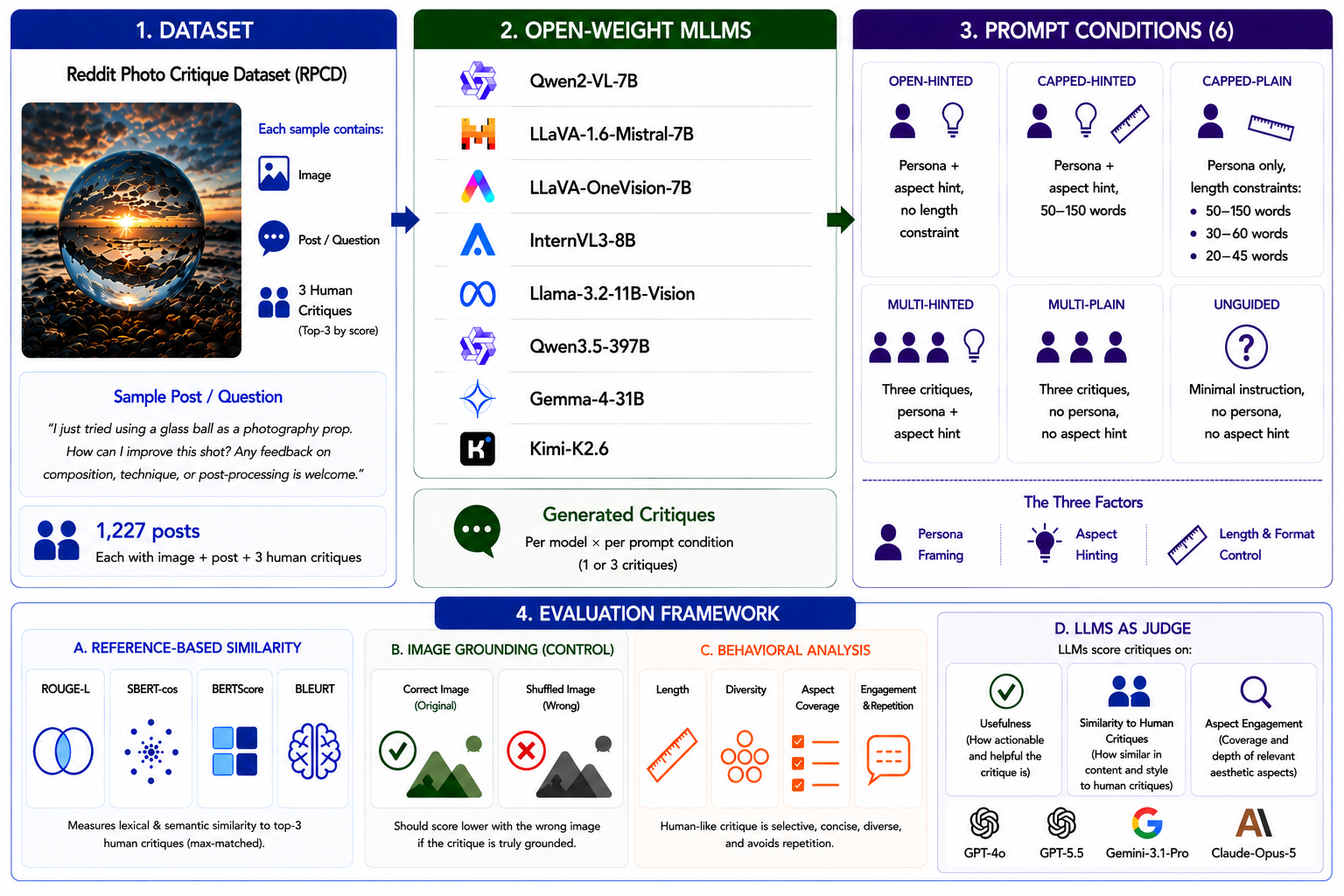}
\caption{Overview of our evaluation pipeline: $1{,}227$
\texttt{r/photocritique} posts, each with an image and three ranked human
critiques (sample illustrative); eight open-weight MLLMs critiquing every post under eight
prompt conditions; and four evaluation axes, reference-based similarity,
grounding controls, LLM judges, and behavioral comparison.}
\label{fig:info}
\end{figure*}


When a photographer posts an image to an online critique community and asks
for feedback, what they receive back is not a number. It is a short paragraph:
a few targeted observations about what works, what does not, and what to try
next. Computational work on image aesthetics, however, has historically
reduced this act to \emph{scoring}. Benchmarks such as AVA~\cite{murray2012ava},
predictors such as NIMA~\cite{talebi2018nima}, and attribute-level
datasets~\cite{kong2016photo,kang2020eva,fang2020perceptual} pair each image
with a well-defined numeric target, so that model quality reduces to a
correlation against a ground-truth score. Aesthetic \emph{critique} is the
form photographic feedback actually takes, and being textual, open-ended, and
multiply-valued, it is precisely the form that resists such evaluation.

Instruction-tuned multimodal large language models (MLLMs) have made such
critique trivially easy to elicit. Open-weight families such as
Qwen2-VL~\cite{wang2024qwen2vl}, LLaVA~\cite{liu2024llavanext,li2024llavaonevision},
InternVL3~\cite{zhu2025internvl3}, and
Llama-3.2-Vision~\cite{grattafiori2024llama3}, and frontier systems such as
Qwen3.5~\cite{qwen35}, Gemma~4~\cite{gemma4}, Kimi-K2.6~\cite{kimi26}, and
GPT-5.5~\cite{gpt55}, all produce fluent, structured commentary about an image
from a single prompt, and a growing body
of work repurposes them as aesthetic raters and
explainers~\cite{wu2024q,huang2024aesbench,zhou2024uniaa}, with several reports
suggesting they approach or even surpass human references on
aesthetic-\emph{assessment} tasks~\cite{zhou2024uniaa}. This invites a natural
question, and the one we ask here: when an MLLM critiques a photograph, is it
critiquing like a human, or merely producing text that \emph{scores} as if it
were?

The question is difficult because open-ended critique has no single correct
response and is conventionally scored against human references using
similarity metrics~\cite{lin2004rouge,zhang2019bertscore,sellam2020bleurt},
which are convenient and reproducible but correlate only weakly with human
judgment when many distinct responses are equally
acceptable~\cite{novikova2017we,mathur2020tangled}: a critique can be
excellent yet share little surface form with any reference, or mediocre yet
score well by echoing reference vocabulary. Work on open-ended social
reasoning reaches the same conclusion outside the visual domain, scoring
responses to \texttt{r/ChangeMyView} prompts against multiple human
references with a max-mapping and finding that aggregate similarity obscures
persistent human--model gaps~\cite{amirizaniani2024can}. A high similarity
score is not by itself evidence of human-level critique. The now-standard
alternative, asking a strong LLM to judge~\cite{zheng2023judging}, carries its
own verbosity and self-preference
biases~\cite{dubois2024length,panickssery2024llm}, so it too must be tested
rather than trusted.

We therefore evaluate eight open-weight MLLMs from $7$B to $397$B, plus
GPT-5.5 as a closed-weight reference point, against ranked human references
drawn from \texttt{r/photocritique}~\cite{veranieto2022understanding}, which
supplies several independently upvoted critiques per photo. This multiplicity
is central to our design: competent critiques of the same image routinely
disagree, so we adopt a charitable multiple-reference, max-mapping protocol
that rewards a model critique for resembling \emph{any} human reference rather
than an arbitrary one, together with a leave-one-out human agreement baseline
that calibrates the range competent human critiques occupy under each metric.
To separate how a critique is elicited from how a model critiques, we vary
persona framing, aspect hinting, length control, and single- versus
multi-pass generation across eight prompt conditions, two of them at the human
word count. To test whether scores reflect engagement with the photograph
rather than a generic critiquing register, we add grounding controls that keep
the correct title and body but give the model the \emph{wrong} image or none
at all. And we put judging to the same test, having four LLM judges rate critique pairs, validated against two human annotators.
Fig.~\ref{fig:info} summarizes the pipeline.


Across these analyses we find that reference-based similarity gives a
misleading picture of MLLM critique. The stricter lexical and learned metrics
show only weak alignment between model and human critiques, yet a coarse
embedding cosine reports broad topical overlap, so read naively these scores
can suggest parity with human critique. Two controls show why that reading is
mistaken. Asking for shorter critiques, with the model and the images
unchanged, raises the same scores substantially, so they track length as much
as content; and withholding the image changes the critiques surprisingly
little, so much of the measured similarity comes from the shared post text and
a stable critiquing style rather than from what the model saw. An LLM judge sharpens the
question rather than settling it. In the primary condition all four judges
prefer the frontier models' critiques to the matched human ones, but for the
$7$--$8$B models their verdicts diverge wildly, from $9\%$ to $81\%$
preference on identical pairs. What
they do agree on, and what our two human annotators agree with them about, is
that a model critique and a human one are rarely saying the same thing: mean
substantive-similarity ratings run from $1.81$ to $2.59$ on a $1$--$5$ scale,
around the ``mostly different'' mark. The behavioral evidence is more consistent. Models cover nearly
every aesthetic aspect where humans are selective and repeat themselves across
critiques of the same photo where humans vary, even when the prompt caps
length in the human range. Qualitatively, smaller models walk
through a fixed set of aspects rather than answer the question asked, reframe
likely flaws as intentional choices, and stay uniformly positive where human
critics diagnose problems and suggest fixes (Appendix \ref{sec:app-examples}); frontier models answer the
question and name the defect, but keep the same comprehensive register. We
read these gaps as a measurement artifact, not as evidence that these models
critique like humans.


\noindent\textbf{Contributions.} \textbf{(i)}~We treat open-ended aesthetic
critique as an evaluation problem in its own right, and build a benchmark of
$1{,}227$ \texttt{r/photocritique} posts in which every photo carries three
ranked critiques written by someone other than the poster, scored under a
charitable multiple-reference max-mapping protocol calibrated by a
leave-one-out human agreement baseline. \textbf{(ii)}~We evaluate eight
open-weight MLLMs from $7$B to $397$B, plus GPT-5.5, under eight prompt
conditions, two of them length-matched to human critiques, with
shuffled-image and text-only controls that isolate what the photograph
contributes. \textbf{(iii)}~We subject the mainstream alternative to the same
scrutiny, having four LLM judges rate both pairwise usefulness and substantive
similarity on identical critique pairs, and validating them against two human
annotators. \textbf{(iv)}~We characterize how MLLM critique behaves along
length, diversity, aspect coverage, and engagement, and analyze qualitatively
how that behavior changes across the capability range.


\section{Related Work}

\textbf{Aesthetic assessment and critique datasets.} Computational work on
image aesthetics has historically centered on \emph{scoring}.
AVA~\cite{murray2012ava} established score regression and binary high/low
classification as the dominant formulation, predictors such as
NIMA~\cite{talebi2018nima} learned to reproduce the human score
distribution, and attribute-level datasets (AADB~\cite{kong2016photo},
EVA~\cite{kang2020eva}, SPAQ~\cite{fang2020perceptual}) refined the target
while keeping it numeric, so model quality reduces to a correlation against a
ground-truth score. Aesthetic \emph{critique} is instead textual,
open-ended, and multiply-valued.
PCCD~\cite{chang2017aesthetic} introduced aspect-centric written critiques,
and AVA-Comments and AVA-Captions~\cite{ghosal2019aesthetic} scaled critique
up by mining community comments. The Reddit Photo Critique
Dataset~\cite{veranieto2022understanding} draws critiques from
\texttt{r/photocritique} and, crucially for our purposes, supplies several
independently \emph{ranked} critiques per photo: competent critiques of the
same photo disagree, and any sensible evaluation must contend with that
spread rather than treat one as canonical.

\textbf{Multimodal LLMs as aesthetic critics.} Instruction-tuned multimodal
large language models turn vision--language
pretraining~\cite{radford2021learning,li2023blip} into fluent, on-demand
commentary. Open-weight families such as Qwen2-VL~\cite{wang2024qwen2vl},
LLaVA-NeXT~\cite{liu2024llavanext},
LLaVA-OneVision~\cite{li2024llavaonevision},
InternVL3~\cite{zhu2025internvl3}, and
Llama-3.2-Vision~\cite{grattafiori2024llama3}, like proprietary systems such
as GPT-4V~\cite{openai2023gpt4v}, write a structured paragraph about an image
from a single prompt, and a growing body of work repurposes them as
aesthetic raters and explainers: Q-Bench~\cite{wu2024q},
AesBench~\cite{huang2024aesbench}, and UNIAA~\cite{zhou2024uniaa} probe
aesthetic and low-level perception, Q-Align~\cite{wu2023q} adapts them for
scoring, The Photographer's Eye~\cite{qi2025photographer} trains them to
critique like photographers, and related work examines bias and
personalization~\cite{li2025aesbiasbench} or efficient
adaptation~\cite{ghiasvand2026mmlop, ghiasvand2025pfedmma}. Several report
MLLMs approaching or surpassing human references on
aesthetic-\emph{assessment} tasks~\cite{zhou2024uniaa}, but those are
numeric-scoring results. On open-ended critique we find that apparent parity
coexists with divergence in length, selectivity, and diversity,
and that the disagreement between metrics widens with model scale rather
than resolving.


\textbf{Evaluating open-ended generation.} Reference-based metrics from
machine translation and summarization, from n-gram overlap
(BLEU~\cite{papineni2002bleu}, ROUGE~\cite{lin2004rouge}) to embedding- and
learning-based measures (BERTScore~\cite{zhang2019bertscore},
BLEURT~\cite{sellam2020bleurt}, MoverScore~\cite{zhao2019moverscore}), are
reproducible but correlate weakly with human judgment on multiply-valued
outputs~\cite{novikova2017we,mathur2020tangled}: a generation can be
excellent yet share little surface form with any reference, or mediocre yet
score well by echoing reference vocabulary. Work on open-ended social
reasoning makes the same point outside the visual
domain~\cite{amirizaniani2024can}, scoring responses to Reddit
\texttt{r/ChangeMyView} prompts against multiple human references with a
max-mapping and finding that aggregate similarity obscures human--model
gaps. We adopt a comparable protocol for visual critique and add
prompt-condition, shuffled-image, and text-only controls.


\textbf{LLMs as evaluators.} A complementary line of work therefore uses a
strong LLM as a judge, scoring outputs against a rubric or expressing
pairwise preferences~\cite{zheng2023judging,liu2023g,aryashad2026filters}.
Judges track humans better than n-gram metrics in many settings but exhibit
position and verbosity biases~\cite{wang2024large,dubois2024length} and a
self-preference for text from their own
lineage~\cite{panickssery2024llm}. As judging is now the mainstream
alternative for open-ended outputs, we treat it as an object of study rather
than an oracle: four judges rate identical critique pairs, and on the smaller
models their verdicts range from $9\%$ to $81\%$ preference for the model,
while their similarity ratings agree closely with one another and with two
human annotators.

\section{Experimental Setup}
\label{sec:setup}



\textbf{Dataset.}
Our evaluation set is drawn from \texttt{r/photocritique}, where
photographers post a photograph with a short description and receive
written critiques from the community. The subreddit suits our setting
because it supplies several independently up-voted critiques per photo, and
this multiplicity is central to our analysis: competent critiques of the
same image routinely disagree, and any faithful evaluation of open-ended
critique must measure against that spread rather than treat a single
critique as canonical. As the candidate pool we take the released
test-split identifiers of the Reddit Photo Critique Dataset
(RPCD)~\cite{veranieto2022understanding}, retrieving each submission,
image, and comment thread directly from public Reddit archives rather than
from the dataset's distributed text.

A post is kept only if (i)~it has an accessible image, (ii)~its title and
body together contain at least ten words, ensuring the model is conditioned
on a substantive prompt, and (iii)~it has at least three \emph{qualifying}
comments: top-level critiques of at least five words that are not written
by the submitting user,\footnote{\texttt{r/photocritique} requires posters
to leave a context comment under their own submission; these describe the
shoot rather than critique it. Posts whose submitting account has been
deleted are excluded, since comment authorship can no longer be verified.}
not exact duplicates of another comment, and not moderator notices. For
each surviving post we rank the qualifying comments by community score and
keep the top three as human references. The resulting benchmark contains
$1{,}227$ posts, each paired with a verified image and exactly three ranked
human critiques; unless noted otherwise, all reported scores are averaged
over these posts.


\textbf{Models.}
Our focus is open-weight MLLMs, and we evaluate eight of them spanning the
capability range: five models of comparable scale run locally,
Qwen2-VL-7B-Instruct~\cite{wang2024qwen2vl}, LLaVA-1.6-Mistral-7B
(LLaVA-NeXT)~\cite{liu2024llavanext},
LLaVA-OneVision-7B~\cite{li2024llavaonevision},
InternVL3-8B~\cite{zhu2025internvl3}, and
Llama-3.2-11B-Vision-Instruct~\cite{grattafiori2024llama3}, and three
frontier-class models included to test whether our findings are an artifact
of model scale, Qwen3.5-397B-A17B~\cite{qwen35}, Gemma-4-31B~\cite{gemma4},
and Kimi-K2.6~\cite{kimi26}, served with the same prompts and
full-resolution images. As a closed-weight reference
point we additionally evaluate GPT-5.5~\cite{gpt55}; Kimi-K2.6 and GPT-5.5
are run in the primary condition only. All models share identical decoding settings, nucleus
sampling with temperature $0.5$, top-$p\,{=}\,0.9$, and a fixed seed of
$42$, so that variation across models and conditions reflects the prompt and
the model rather than the sampler; the exception is GPT-5.5, which does not
expose sampling controls. The only generation parameter that changes is the
token budget \texttt{max\_new\_tokens}, set per condition to accommodate the
requested output length.\footnote{\texttt{max\_new\_tokens} is set per
condition to match the requested output: $200$ for the capped
single-critique conditions (\textsc{Capped-Plain}, \textsc{Capped-Hinted}),
$512$ for the uncapped single-critique conditions (\textsc{Open-Hinted},
\textsc{Unguided}), $600$ for the multi-critique conditions
(\textsc{Multi-Plain}, \textsc{Multi-Hinted}), and $160$ and $130$ for the
length-matched \textsc{Capped-Plain} variants. All other decoding parameters
are shared.} Under every condition each model produces three critiques per
post, mirroring the three human references and enabling a symmetric
model-vs-human comparison.

\begin{figure}[t]
\centering

\begin{promptbox}{Prompt Template --- \textsc{Capped-Plain}}
You are an experienced photographer browsing r/photocritique, an online community
where photographers share their work and receive constructive feedback.

\smallskip
A user posted this photo with the title: ``\pv{title}'' \pv{body}

\smallskip
Provide a focused critique in 50--150 words. Be specific and constructive.
\end{promptbox}
\caption{Our primary \textsc{Capped-Plain} prompt: persona framing with a 50--150 word
constraint and no aspect hint. }
\label{fig:prompt-capped-plain}
\end{figure}
\textbf{Prompt conditions.}
To separate \emph{how} a critique is elicited from a model's underlying
critiquing behavior, we evaluate every model under six prompt conditions
that vary three factors: \emph{persona framing} (whether the
model is cast as an experienced photographer), \emph{aspect hinting}
(whether the prompt enumerates aesthetic aspects to consider), and
\emph{length and format control} (whether output length is capped, and
whether one or three critiques are requested in a single pass). The six
conditions are:
\begin{itemize}
  \item \textsc{Open-Hinted}: persona framing and an aspect hint, with no
        length constraint (Fig.~A1 in the Appendix).
  \item \textsc{Capped-Hinted}: \textsc{Open-Hinted} with an added
        50--150 word constraint (Fig.~A2 in the Appendix).
  \item \textsc{Capped-Plain}: persona framing and a 50--150 word
        constraint, with the aspect hint removed
        (Fig.~\ref{fig:prompt-capped-plain}); our primary condition.
  \item \textsc{Multi-Hinted}: a single pass producing three critiques,
        each assigned a different angle, with persona and aspect hints
        (Fig.~A3 in the Appendix).
  \item \textsc{Multi-Plain}: a single pass producing three critiques as
        post comments, with no persona and no aspect hint
        (Fig.~A4 in the Appendix).
  \item \textsc{Unguided}: a minimal instruction with no persona, no aspect
        hint, and no length constraint (Fig.~A5).
\end{itemize}

We designate \textsc{Capped-Plain} as the primary condition because it
constrains length to a human-plausible range while leaving the model free to
choose what to discuss, isolating critiquing behavior from prompt-supplied
scaffolding. We add two length-matched variants of it, identical
except that the $50$--$150$ words become $30$--$60$ and $20$--$45$
to bracket the human mean of $54$ words;
Section~\ref{sec:results} uses them to test whether the reported differences
survive at human-comparable length. Every
range is a natural-language instruction rather than hard truncation;
Appendix~\ref{sec:app-length} reports how well each model complies and where
the decoding budget binds instead. The full
prompt templates are given in Figure~\ref{fig:prompt-capped-plain} and
Figs.~\ref{fig:prompt-open-hinted}--\ref{fig:prompt-unguided} in the
Appendix.

\textbf{Similarity Metrics.} We score each model critique against the human references with four reference-based metrics, chosen to span lexical, embedding-level, and learned-semantic notions of similarity:

\begin{itemize}
    \item \textbf{ROUGE-L} \cite{lin2004rouge}: standard longest-common-
    subsequence overlap (stemmed), capturing lexical agreement.
    \item \textbf{SBERT-cos}: cosine similarity between sentence embeddings of
    the model and human critiques, computed with the
    \texttt{all-mpnet-base-v2} sentence-transformer~\cite{reimers2019sentence}
    (mean pooling over token embeddings, each critique encoded whole). Each
    critique becomes a single fixed-length vector, so the cosine gives a coarse
    measure of topical agreement, insensitive to exact wording.
    \item \textbf{BERTScore} \cite{zhang2019bertscore}: token-level greedy matching over contextual embeddings from RoBERTa-large
    \cite{liu2019roberta}, with baseline rescaling. Rescaling subtracts the
    score expected for unrelated text in the same language, so values near zero
    indicate token-level overlap comparable to that of unrelated critiques; the
    rescaled scale is a readability convention, not a calibrated probability.
    \item \textbf{BLEURT} \cite{sellam2020bleurt}: the learned BLEURT-20
    checkpoint. As BLEURT is trained to judge translation-grade equivalence
    rather than loose topical relatedness, negative scores are expected here
    and their absolute magnitude carries no calibrated meaning for this task;
    we use BLEURT only to compare critiques with one another.
\end{itemize}

We include metrics at three levels of strictness because they
disagree in informative ways: SBERT-cos rewards broad topical overlap, whereas
ROUGE-L, BERTScore, and BLEURT each demand progressively closer correspondence
in wording or meaning. As we show in Section~\ref{sec:results}, this
disagreement is itself one of our central findings.

\textbf{Aggregation: 3$\times$3 Max-Mapping.} Each post has three model
critiques and three human references with no canonical pairing between them.
For every metric we compute the $3\times3$ similarity matrix, take the
row-wise maximum, and average across rows. This \emph{max-mapping} score
rewards a model critique for resembling \emph{any} human reference rather than
an arbitrary one, the most charitable reasonable aggregation given the spread
among human critiques.

To calibrate its range we compute a \textbf{human agreement baseline} the same
way, leave-one-out: each human critique in turn is the candidate and the
remaining two the references. It measures how similar one competent critique
is to the others of the same photo, and gives the scores a scale. It is not a
ceiling: exceeding it means only that a critique resembles the reference set
more closely than the references resemble one another, which a fluent,
comprehensive critique can do without being more human-like. Nor is the
comparison perfectly symmetric, since a leave-one-out candidate consumes one
of its own references and row-wise maximization lets several model critiques
claim the same one; Appendices~\ref{sec:aggregation-robustness}
and~\ref{sec:ceiling-sensitivity} re-derive every score under stricter
aggregations and a debiased baseline, and no conclusion depends on the choice.

\section{Results}
\label{sec:results}

\begin{table}[t]
\centering
\caption{Critique--human similarity under the primary \textsc{Capped-Plain} condition ($N{=}1{,}227$, 3$\times$3 max-mapping). Bold marks the best model per metric. }
\label{tab:main}
\setlength{\tabcolsep}{2.5pt}
\resizebox{0.4\textwidth}{!}{%
\small
\begin{tabular}{lccccc}
\toprule
Model & Words & ROUGE-L & SBERT-cos & BERTScore & BLEURT \\
\midrule
Qwen2-VL-7B          & 112.4 & 0.142 & 0.621 & 0.0986 & $\mathbf{-0.601}$ \\
LLaVA-1.6-Mistral-7B & 150.1 & 0.134 & 0.594 & 0.0548 & $-0.698$ \\
LLaVA-OneVision-7B   & 143.0 & 0.132 & 0.611 & 0.0655 & $-0.631$ \\
InternVL3-8B         & 129.2 & 0.128 & 0.615 & 0.0736 & $-0.602$ \\
Llama-3.2-11B-Vision & 150.7 & 0.141 & 0.626 & 0.0834 & $-0.619$ \\
Qwen3.5-397B-A17B    & 118.8 & 0.135 & 0.641 & 0.0816 & $-0.852$ \\
Gemma-4-31B          & 121.5 & \textbf{0.147} & 0.648 & \textbf{0.1031} & $-0.876$ \\
Kimi-K2.6            & 101.0 & 0.139 & 0.632 & 0.0846 & $-0.932$ \\
GPT-5.5              & 115.9 & 0.143 & \textbf{0.649} & 0.0986 & $-0.880$ \\
\midrule
\textit{Human agreement} &  54.0 & 0.147 & 0.514 & 0.1139 & $-1.077$ \\
\bottomrule
\end{tabular}%
}
\end{table}


\begin{figure*}[t]
\centering
\includegraphics[width=0.9\textwidth]{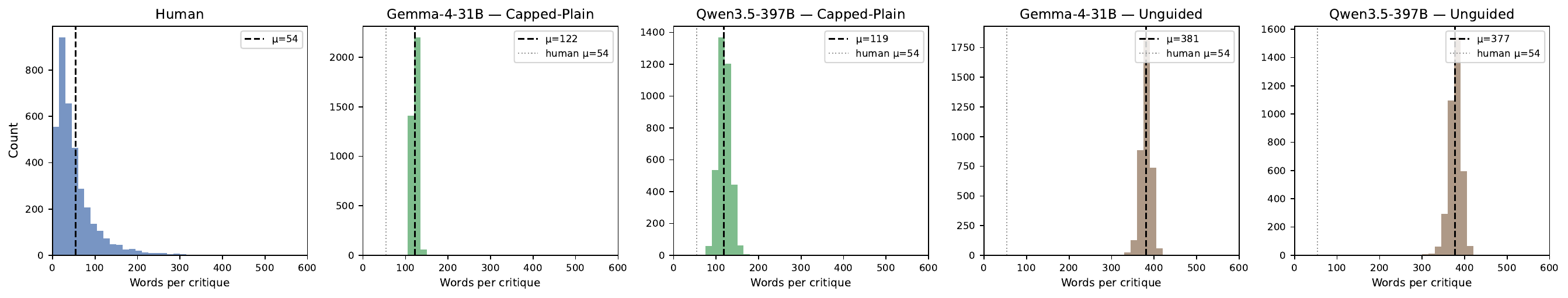}
\caption{Critique length for Gemma-4-31B and Qwen3.5-397B under \textsc{Capped-Plain} and \textsc{Unguided} conditions, with the human distribution at left ($N{=}1{,}227$); dashed lines mark each mean. Per-model distributions for all conditions are in Figs.~\ref{fig:length-hist} and~\ref{fig:length-hist-capped}.}
\label{fig:hist-main}
\end{figure*}


\begin{table}[t]
\centering
\caption{Similarity by prompt condition for the two frontier-class open-weight
models with full condition coverage, Qwen3.5-397B and Gemma-4-31B
($N{=}1{,}227$). \emph{Human} is the leave-one-out agreement baseline.
Remaining models in Tables~\ref{tab:full-similarity}
and~\ref{tab:length-matched}.}
\label{tab:conditions}
\setlength{\tabcolsep}{2.5pt}
\resizebox{0.4\textwidth}{!}{%
\small
\begin{tabular}{lccccc}
\toprule
Condition & Words & R-L & SBERT-cos & BERTScore & BLEURT \\
\midrule
\multicolumn{6}{l}{\textit{Qwen3.5-397B}} \\
Unguided             & 376.9 & 0.096 & 0.635 & $-0.0824$ & $-0.924$ \\
Open-Hinted          & 396.5 & 0.095 & 0.644 & $-0.0510$ & $-0.927$ \\
Capped-Hinted        & 134.7 & 0.131 & 0.634 & $0.0684$  & $-0.793$ \\
Capped-Plain         & 118.8 & 0.135 & 0.641 & $0.0816$  & $-0.852$ \\
Capped-Plain (30--60) &  47.6 & 0.146 & 0.624 & $0.1346$  & $-0.894$ \\
Capped-Plain (20--45) &  40.3 & 0.146 & 0.613 & $0.1443$  & $-0.883$ \\
Multi-Hinted         &  98.6 & 0.136 & 0.582 & $0.0744$  & $-0.810$ \\
Multi-Plain          &  94.5 & 0.138 & 0.633 & $0.0960$  & $-0.856$ \\
\midrule
\multicolumn{6}{l}{\textit{Gemma-4-31B}} \\
Unguided             & 380.5 & 0.101 & 0.646 & $-0.0731$ & $-0.937$ \\
Open-Hinted          & 375.5 & 0.104 & 0.645 & $-0.0275$ & $-0.945$ \\
Capped-Hinted        & 125.2 & 0.146 & 0.648 & $0.1040$  & $-0.837$ \\
Capped-Plain         & 121.5 & 0.147 & 0.648 & $0.1031$  & $-0.876$ \\
Capped-Plain (30--60) &  49.4 & 0.163 & 0.631 & $0.1638$  & $-0.822$ \\
Capped-Plain (20--45) &  37.1 & 0.163 & 0.618 & $0.1751$  & $-0.833$ \\
Multi-Hinted         &  99.6 & 0.149 & 0.586 & $0.1020$  & $-0.739$ \\
Multi-Plain          &  94.6 & 0.159 & 0.638 & $0.1376$  & $-0.745$ \\
\midrule
\textit{Human} & 54.0 & 0.147 & 0.514 & $0.1139$ & $-1.077$ \\
\bottomrule
\end{tabular}
}
\end{table}

\textbf{Reference-Based Similarity.}
Table~\ref{tab:main} reports critique--human similarity under the primary
\textsc{Capped-Plain} condition for all nine models, from the 7--11B
open-weight models to the frontier-scale Kimi-K2.6 and the closed-weight
GPT-5.5. The most immediate asymmetry is length. Human critiques
average only $54.0$ words, whereas every model writes two to three times as
much ($101$--$151$ words), and this is \emph{under} a 50--150 word cap;
unconstrained they run far longer still. The gap matters because length, as we show
below, drives much of the similarity these metrics report.


\begin{table}[t]
\centering
\caption{Image-grounding test (SBERT-cos; $N{=}500$ posts). \textsc{C-vs-H}, \textsc{S-vs-H} and \textsc{T-vs-H} score critiques written with the correct image, a shuffled image and no image against the same human references; the largest of the three is bold. $\Delta_{\mathrm{S}}$ and $\Delta_{\mathrm{T}}$ give the advantage of the correct image over each alternative. Kimi-K2.6 and GPT-5.5 were run in the primary condition only.}
\label{tab:grounding}
\setlength{\tabcolsep}{2.5pt}
\resizebox{0.38\textwidth}{!}{%
\small
\begin{tabular}{lccccc}
\toprule
Model & C-vs-H & S-vs-H & T-vs-H & $\Delta_{\mathrm{S}}$ & $\Delta_{\mathrm{T}}$ \\
\midrule
\multicolumn{6}{l}{\textit{Open-Hinted}} \\
\midrule
Qwen2-VL-7B          & $\mathbf{0.614}$ & $0.498$ & $0.570$ & $+0.116$ & $+0.044$ \\
LLaVA-1.6-Mistral-7B & $0.591$ & $0.512$ & $\mathbf{0.600}$ & $+0.079$ & $-0.009$ \\
LLaVA-OneVision-7B   & $\mathbf{0.606}$ & $0.498$ & $0.575$ & $+0.107$ & $+0.030$ \\
InternVL3-8B         & $\mathbf{0.614}$ & $0.494$ & $0.564$ & $+0.120$ & $+0.050$ \\
Llama-3.2-11B-Vision & $\mathbf{0.612}$ & $0.522$ & $0.589$ & $+0.090$ & $+0.023$ \\
Qwen3.5-397B         & $\mathbf{0.644}$ & $0.529$ & $0.584$ & $+0.115$ & $+0.060$ \\
Gemma-4-31B          & $\mathbf{0.648}$ & $0.534$ & $0.595$ & $+0.113$ & $+0.052$ \\
\midrule
\multicolumn{6}{l}{\textit{Capped-Hinted}} \\
\midrule
Qwen2-VL-7B          & $\mathbf{0.597}$ & $0.482$ & $0.577$ & $+0.115$ & $+0.019$ \\
LLaVA-1.6-Mistral-7B & $0.562$ & $0.479$ & $\mathbf{0.599}$ & $+0.083$ & $-0.037$ \\
LLaVA-OneVision-7B   & $\mathbf{0.580}$ & $0.475$ & $0.566$ & $+0.105$ & $+0.014$ \\
InternVL3-8B         & $\mathbf{0.593}$ & $0.478$ & $0.569$ & $+0.114$ & $+0.024$ \\
Llama-3.2-11B-Vision & $0.598$ & $0.507$ & $\mathbf{0.603}$ & $+0.091$ & $-0.005$ \\
Qwen3.5-397B         & $\mathbf{0.634}$ & $0.518$ & $0.594$ & $+0.116$ & $+0.040$ \\
Gemma-4-31B          & $\mathbf{0.650}$ & $0.543$ & $0.597$ & $+0.107$ & $+0.053$ \\
\midrule
\multicolumn{6}{l}{\textit{Capped-Plain}} \\
\midrule
Qwen2-VL-7B          & $\mathbf{0.624}$ & $0.516$ & $0.600$ & $+0.108$ & $+0.024$ \\
LLaVA-1.6-Mistral-7B & $0.598$ & $0.513$ & $\mathbf{0.603}$ & $+0.085$ & $-0.005$ \\
LLaVA-OneVision-7B   & $\mathbf{0.612}$ & $0.514$ & $0.588$ & $+0.097$ & $+0.023$ \\
InternVL3-8B         & $\mathbf{0.616}$ & $0.503$ & $0.584$ & $+0.113$ & $+0.032$ \\
Llama-3.2-11B-Vision & $\mathbf{0.628}$ & $0.533$ & $0.608$ & $+0.094$ & $+0.020$ \\
Qwen3.5-397B         & $\mathbf{0.643}$ & $0.531$ & $0.598$ & $+0.112$ & $+0.046$ \\
Gemma-4-31B          & $\mathbf{0.649}$ & $0.544$ & $0.579$ & $+0.105$ & $+0.069$ \\
Kimi-K2.6            & $\mathbf{0.634}$ & $0.527$ & $0.541$ & $+0.107$ & $+0.094$ \\
GPT-5.5              & $\mathbf{0.651}$ & $0.541$ & $0.579$ & $+0.110$ & $+0.072$ \\
\bottomrule
\end{tabular}%
}
\end{table}

Turning to the scores themselves, three of the four metrics agree that
content overlap with human critiques is weak in absolute terms: ROUGE-L sits
at $0.13$--$0.15$ for every model, BERTScore hovers near zero, and BLEURT is
strongly negative. On these measures, model critiques rarely reproduce what
humans actually write, and direct ratings agree: human annotators and four
LLM judges rating substantive similarity place model critiques between
``mostly different'' and ``half similar'' on a 1--5 scale
(Tables~\ref{tab:human-validation} and~\ref{tab:sim-judge}). SBERT-cos is the lone
dissenter at roughly $0.6$ for all models, exactly what a coarse embedding
cosine returns for broad topical overlap rather than substantive agreement.
Read against the human agreement baseline, the metrics split two against
two: all nine models exceed it on SBERT-cos and BLEURT yet fall at or below
it on ROUGE-L and BERTScore. Taken at face value, half the metrics place the
models at or above human agreement, a reading we argue is a measurement
artifact rather than a capability. Importantly, this split is not caused by
the aggregation or by the baseline estimate itself, as we verify in
Secs.~\ref{sec:aggregation-robustness} and~\ref{sec:ceiling-sensitivity}.

Table~\ref{tab:conditions} shows that this picture is not an artifact of one
prompt, reporting the sweep for the two frontier-class open-weight models
with full condition coverage, Qwen3.5-397B and Gemma-4-31B; the remaining
models behave alike (Tables~\ref{tab:full-similarity}
and~\ref{tab:length-matched}). Length itself swings widely with the prompt,
from $397$ words under \textsc{Open-Hinted} down to $37$ under the tightest
budget, yet every condition that leaves length free far exceeds the human
average. Figure~\ref{fig:hist-main} makes the gap concrete: under the
$50$--$150$ cap both models settle at $119$--$122$ words against a human mean
of $54$, and uncapped they run to $377$--$381$, with nearly the entire model
distribution lying to the right of the human one (per-model distributions in
Figs.~\ref{fig:length-hist} and~\ref{fig:length-hist-capped}). Each metric
mostly tracks this surface property rather than quality. For Gemma, ROUGE-L
rises monotonically as critiques shorten, from $0.101$ under
\textsc{Unguided} to $0.163$ under the length-matched budgets, because
shorter text dilutes the reference vocabulary less, and BERTScore climbs from
negative to $0.175$: the tightened word range alone pushes Gemma past the
human baseline on exactly the two metrics on which models otherwise fall
short.
SBERT-cos is nearly prompt-invariant, holding between $0.61$ and $0.65$
except under \textsc{Multi-Hinted}. No condition optimizes all four metrics
at once, and which prompt looks best depends on which metric is
reported.

\textbf{Image-Grounding Test.}
A natural worry is that these scores reflect a generic critiquing register
rather than genuine engagement with the photograph.  We test this with two
controls that keep the correct title and body and vary only what the model
sees: the \emph{wrong} (shuffled) image, or no image at all
(Table~\ref{tab:grounding}). Three comparisons are relevant:
correct-image critique vs.\ human references (C-vs-H), shuffled-image
critique vs.\ the correct image's references (S-vs-H), and text-only
critique, written without any image, vs.\ those same references (T-vs-H).
$\Delta_{\mathrm{S}} = \text{C-vs-H} - \text{S-vs-H}$ and
$\Delta_{\mathrm{T}} = \text{C-vs-H} - \text{T-vs-H}$ measure the advantage
of seeing the correct image over the shuffled image and over the text alone.
$\Delta_{\mathrm{S}}$ is positive in every cell, averaging $+0.104$, so
grounding is real. Its magnitude, however, is modest: shuffled-image
critiques still reach $0.475$--$0.544$, meaning a model describing a photo
it never saw matches the human references almost as closely as one that did.

The text-only control asks how much of that residual similarity is explained
by conditioning on the shared post text. The answer is most of it. Text-only
critiques score $0.541$--$0.608$ against the human references, and the
advantage of the correct photograph over the text alone averages just
$+0.03$, less than a third of the shuffled-image gap. We therefore attribute
the high SBERT-cos scores in Table~\ref{tab:main} to shared textual
conditioning combined with a stable critiquing style, rather than to
image-specific observation. Two further patterns stand out. First,
$\Delta_{\mathrm{T}}$ tracks model capability, running from $-0.037$ for
LLaVA-1.6-Mistral-7B, which scores \emph{higher} without the image, to
$+0.094$ for Kimi-K2.6\footnote{Kimi-K2.6 also declined to critique without
an image on $11.0\%$ of posts, against $0.2$--$1.2\%$ for every other
model.}; the two largest values in the study belong to the two frontier
models, the open-weight Kimi-K2.6 and the closed-weight GPT-5.5
($+0.072$), so the weak grounding elsewhere reflects the models rather than
the task. Second, S-vs-H falls below T-vs-H in every row: a wrong
photograph is worse than none, so shuffled-image controls alone understate
how much of the similarity the text explains. The same pattern
holds under the stricter BERTScore (Table~\ref{tab:grounding-bert}), where
$\Delta_{\mathrm{S}}$ stays positive in $22$ of $23$ cells and the frontier
models again sit at the image-dependent end.

\begin{table}[t]
\centering
\caption{Pairwise usefulness: share (\%) of pairs in which each judge
preferred the MLLM critique over the matched human critique, under two
output-length budgets ($100$ posts per cell). Judges: GPT-4o
(\jicon{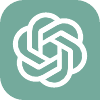}\,4o), GPT-5.5 (\jicon{openai}\,5.5), Gemini-3.1-Pro
(\jicon{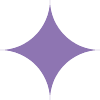}), Claude-Opus-5 (\jicon{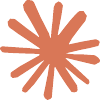}); \textit{Mean}
averages the four.}
\label{tab:usefulness-pref}
\setlength{\tabcolsep}{3pt}
\resizebox{0.45\textwidth}{!}{%
\small
\begin{tabular}{lccccc ccccc}
\toprule
& \multicolumn{5}{c}{Capped-Plain (50--150)} & \multicolumn{5}{c}{Capped-Plain (30--60)} \\
\cmidrule(lr){2-6} \cmidrule(lr){7-11}
Model & \jicon{openai}\,4o & \jicon{openai}\,5.5 & \jicon{googlegemini} & \jicon{claude} & \textit{Mean} & \jicon{openai}\,4o & \jicon{openai}\,5.5 & \jicon{googlegemini} & \jicon{claude} & \textit{Mean} \\
\midrule
Qwen2-VL-7B  & 73 & 39 & 11 & 11 & \textit{33.5} & 57 & 29 & 15 & 14 & \textit{28.7} \\
InternVL3-8B & 81 & 38 & 12 &  9 & \textit{35.0} & 57 & 30 & 16 &  8 & \textit{27.8} \\
Qwen3.5-397B & 95 & 78 & 73 & 52 & \textit{74.5} & 68 & 63 & 57 & 42 & \textit{57.5} \\
Gemma-4-31B  & 91 & 84 & 79 & 67 & \textit{80.2} & 74 & 63 & 60 & 46 & \textit{60.8} \\
Kimi-K2.6    & 96 & 95 & 94 & 84 & \textit{92.2} & 84 & 87 & 81 & 64 & \textit{79.0} \\
GPT-5.5      & 96 & 99 & 97 & 85 & \textit{94.2} & 86 & 91 & 93 & 75 & \textit{86.2} \\
\bottomrule
\end{tabular}%
}
\end{table}

\begin{table}[t]
\centering
\caption{Human validation of the similarity task: mean $1$--$5$
substantive-similarity rating per model, from two human annotators (pooled)
and from the four LLM judges rating the identical $150$ pairs ($50$ per
model) with the identical rubric (Fig.~\ref{fig:judge-prompt-similarity}).}
\label{tab:human-validation}
\resizebox{0.35\textwidth}{!}{%
\small
\begin{tabular}{lccccc}
\toprule
Model & Human & \jicon{openai}\,4o & \jicon{openai}\,5.5 & \jicon{googlegemini} & \jicon{claude} \\
\midrule
Qwen2-VL-7B  & 2.43 & 2.44 & 2.20 & 1.72 & 1.90 \\
InternVL3-8B & 2.52 & 2.58 & 2.26 & 1.61 & 1.90 \\
Qwen3.5-397B & 2.81 & 2.70 & 2.44 & 2.11 & 2.24 \\
\midrule
\textit{Overall} & \textit{2.59} & \textit{2.57} & \textit{2.30} & \textit{1.81} & \textit{2.01} \\
\bottomrule
\end{tabular}%
}
\end{table}

\begin{table}[t]
\centering
\caption{Aspect coverage (\%) by condition ($N{=}1{,}227$): the fraction of critiques that address each aspect, averaged over posts. \textit{Human} is the reference baseline. Kimi-K2.6 and GPT-5.5 were run in the primary condition only.}
\label{tab:aspect-coverage}
\setlength{\tabcolsep}{2.5pt}
\resizebox{0.45\textwidth}{!}{%
\begin{tabular}{lccccccc}
\toprule
& Comp. & Light. & Color & Expo. & Tech. & Mood & Subj. \\
\midrule
\textit{Human}       & 48.2 & 23.0 & 22.1 & 11.0 & 33.7 & 10.5 & 35.8 \\
\midrule
\multicolumn{8}{l}{\textit{Capped-Plain (50--150)}} \\
\midrule
Qwen2-VL-7B          & 99.6 & 83.3 & 65.5 & 56.0 & 69.8 & 68.1 & 79.3 \\
LLaVA-1.6-Mistral-7B & 99.6 & 81.8 & 69.6 & 27.2 & 65.5 & 73.0 & 82.5 \\
LLaVA-OneVision-7B   & 99.2 & 81.3 & 69.5 & 47.4 & 77.8 & 68.5 & 85.8 \\
InternVL3-8B         & 99.9 & 92.8 & 80.8 & 54.3 & 73.5 & 83.1 & 83.5 \\
Llama-3.2-11B-Vision & 98.4 & 61.6 & 56.5 & 46.1 & 67.1 & 69.0 & 73.3 \\
Qwen3.5-397B         & 98.6 & 75.5 & 75.0 & 34.1 & 73.7 & 82.2 & 77.3 \\
Gemma-4-31B          & 99.2 & 75.5 & 56.9 & 43.2 & 75.1 & 66.5 & 73.2 \\
Kimi-K2.6            & 97.6 & 78.3 & 65.9 & 39.1 & 54.4 & 61.3 & 74.2 \\
GPT-5.5              & 99.0 & 78.3 & 69.1 & 35.7 & 62.9 & 64.4 & 81.6 \\
\midrule
\multicolumn{8}{l}{\textit{Capped-Hinted (50--150)}} \\
\midrule
Qwen2-VL-7B          & 99.9 & 94.7 & 83.8 & 39.4 & 65.8 & 82.2 & 83.1 \\
LLaVA-1.6-Mistral-7B & 99.9 & 97.3 & 77.0 & 11.5 & 42.6 & 82.2 & 82.6 \\
LLaVA-OneVision-7B   & 99.9 & 92.3 & 88.5 & 39.1 & 83.3 & 82.7 & 89.7 \\
InternVL3-8B         & 100.0 & 99.3 & 98.5 & 45.3 & 96.5 & 97.9 & 86.2 \\
Llama-3.2-11B-Vision & 99.8 & 84.7 & 84.6 & 31.7 & 67.1 & 86.7 & 77.0 \\
Qwen3.5-397B         & 99.9 & 85.5 & 92.0 & 31.1 & 84.7 & 95.7 & 81.1 \\
Gemma-4-31B          & 99.8 & 81.1 & 67.4 & 39.7 & 79.4 & 79.2 & 74.9 \\
\midrule
\multicolumn{8}{l}{\textit{Capped-Plain (30--60)}} \\
\midrule
Qwen2-VL-7B          & 95.7 & 68.9 & 56.3 & 42.0 & 50.7 & 46.2 & 65.2 \\
LLaVA-1.6-Mistral-7B & 99.5 & 75.6 & 67.0 & 28.9 & 64.4 & 67.6 & 76.4 \\
LLaVA-OneVision-7B   & 96.9 & 65.7 & 56.0 & 33.9 & 59.9 & 54.4 & 75.1 \\
InternVL3-8B         & 97.4 & 76.0 & 57.3 & 37.6 & 50.7 & 64.3 & 63.9 \\
Llama-3.2-11B-Vision & 91.5 & 39.2 & 28.5 & 36.7 & 49.8 & 38.6 & 51.6 \\
Qwen3.5-397B         & 91.7 & 56.1 & 49.4 & 19.5 & 54.9 & 57.9 & 53.6 \\
Gemma-4-31B          & 94.2 & 56.6 & 36.4 & 27.8 & 51.4 & 41.1 & 48.8 \\
\bottomrule
\end{tabular}%
}
\end{table}

\textbf{LLM-as-Judge Evaluation.}
Since reference-based similarity is unreliable here, we turn to the current
mainstream alternative for open-ended outputs, an LLM
judge~\cite{zheng2023judging}, and test whether its verdict is stable. For
each of $100$ posts and six models, a judge saw the image, the post title
and body, and an anonymized, randomly ordered pair of critiques, one
model-written and one human, and chose the one more useful to the
photographer. Four judges from three model families rated the identical
pairs, GPT-4o~\cite{openai2023gpt4v}, GPT-5.5~\cite{gpt55},
Gemini-3.1-Pro~\cite{gemini31pro} and
Claude-Opus-5~\cite{claudeopus5} (Table~\ref{tab:usefulness-pref}; human and
tie shares in Table~\ref{tab:usefulness-judge}).

The verdict depends as much on the judge as on the critiques: GPT-4o
prefers the two 7--8B models' critiques in $73$--$81\%$ of pairs where
Gemini-3.1-Pro and Claude-Opus-5 prefer them in only $9$--$12\%$, and over
all $600$ pairs the preference for the model critique runs from $51.3\%$
under Claude-Opus-5 to $88.7\%$ under GPT-4o
(Table~\ref{tab:usefulness-pref}). The split does not reflect judge capability: Gemini-3.1-Pro and
Claude-Opus-5, two unrelated judges, agree with each other on $85.8\%$ of
pairs, while GPT-4o agrees with Claude-Opus-5 on barely more than chance
($52.4\%$). The outlier is GPT-4o, whose outputs many open VLMs are
instruction-tuned on, so its favor toward critiques written in its own
style is consistent with known self-preference and verbosity
biases~\cite{dubois2024length,wang2024large,panickssery2024llm}.
Length is part of the preference: a $30$--$60$ word cap lowers every judge
by $6$ to $19$ points. What survives the choice of judge is the tier
structure: all four place every frontier model above both 7--8B models, from
$33$--$35\%$ (mean over judges) for the latter to $94.2\%$
(Table~\ref{tab:usefulness-pref}), and all four prefer the two strongest
models, Kimi-K2.6 and GPT-5.5, over the human critique in a majority of
pairs under both budgets. Within a tier the orderings differ: only GPT-4o
ranks Qwen3.5-397B above Gemma-4-31B. LLM judges agree on capability
tiers, but below the frontier the model-versus-human verdict flips with the
judge.

The disagreement is one of preference, not of perception. Rating instead how
similar each pair is in substance ($1$--$5$), the same judges largely agree
with one another and with the strict metrics: mean pairwise rank correlation
$0.86$, two judges within one point on $98\%$ of pairs, every model near
$2$, and only $13.2\%$ of pairs rated $4$ or above
(Table~\ref{tab:sim-judge} in the Appendix). The judges read the critiques
the same way and diverge only in how they value them.

A human study anchors these ratings (Table~\ref{tab:human-validation}). Two
annotators independently rated $150$ of the same pairs, $50$ each for
Qwen2-VL-7B, InternVL3-8B and Qwen3.5-397B, on the identical scale. They agree with the strict metrics and with the judges' tier ordering:
every model sits between ``mostly different'' and ``half similar'', from
$2.43$ (Qwen2-VL-7B) to $2.81$ (Qwen3.5-397B), and they agree with each
other most about the frontier model (within one point on $85.4\%$ of its
pairs, against $62.0\%$ for Qwen2-VL-7B). Overall agreement is modest, exact
on $31.8\%$ of pairs, within one point on $73.6\%$, Krippendorff's ordinal
$\alpha$ of $0.43$: competent critics disagree about what a photograph
needs, so raters disagree about whether two critiques say the same thing.
Scored the same way, against each annotator in turn and averaged, every
judge tracks the annotators at least as closely as they track each other,
within one point on $73.0$--$85.9\%$ of pairs and $\rho$ of $0.53$--$0.58$
against their $73.6\%$ and $0.43$, while rating $0.1$--$0.8$ points stricter
(Table~\ref{tab:judge-agreement}).

\begin{table}[t]
\centering
\caption{LLM-judge aspect engagement under \textsc{Multi-Plain} ($200$ posts, $600$ critiques per row). GPT-4o rates each critique $0$--$3$ per aspect, from not mentioned to specific observation plus actionable insight (rubric in Fig.~\ref{fig:judge-prompt-aspect}).}
\label{tab:aspect-engagement-judge}
\setlength{\tabcolsep}{2.5pt}
\resizebox{0.45\textwidth}{!}{%
\begin{tabular}{lcccccccc}
\toprule
& Comp. & Light. & Color & Expo. & Tech. & Mood & Subj. & Avg. \\
\midrule
\multicolumn{9}{l}{\textit{Mean engagement depth (0--3)}} \\
\midrule
\textit{Human}       & 1.56 & 0.62 & 0.62 & 0.57 & 0.85 & 0.67 & 1.16 & 0.86 \\
\midrule
Qwen2-VL-7B          & 1.81 & 1.33 & 1.14 & 0.99 & 0.99 & 1.32 & 1.50 & 1.30 \\
LLaVA-1.6-Mistral-7B & 2.18 & 1.36 & 1.13 & 0.65 & 1.22 & 1.53 & 1.73 & 1.40 \\
LLaVA-OneVision-7B   & 2.00 & 1.11 & 1.11 & 0.80 & 1.15 & 1.37 & 1.64 & 1.31 \\
InternVL3-8B         & 2.06 & 1.40 & 1.08 & 0.89 & 1.07 & 1.46 & 1.55 & 1.36 \\
Llama-3.2-11B-Vision & 2.05 & 1.22 & 0.97 & 0.78 & 1.00 & 1.55 & 1.76 & 1.33 \\
Qwen3.5-397B         & 2.66 & 1.49 & 1.56 & 1.23 & 1.50 & 2.01 & 2.02 & 1.78 \\
Gemma-4-31B          & 2.50 & 1.08 & 1.26 & 0.99 & 1.25 & 1.45 & 1.73 & 1.46 \\
\midrule
\multicolumn{9}{l}{\textit{Substantive engagement (\% posts with mean depth $\geq 2$)}} \\
\midrule
\textit{Human}       & 42.5 & 9.5 & 8.0 & 8.5 & 12.5 & 3.0 & 17.5 & 14.5 \\
\midrule
Qwen2-VL-7B          & 47.5 & 25.0 & 17.0 & 10.0 & 16.0 & 22.0 & 33.0 & 24.4 \\
LLaVA-1.6-Mistral-7B & 74.0 & 26.0 & 16.0 & 4.5 & 21.5 & 38.0 & 53.5 & 33.4 \\
LLaVA-OneVision-7B   & 65.0 & 14.0 & 13.5 & 8.5 & 18.5 & 29.5 & 42.5 & 27.4 \\
InternVL3-8B         & 69.5 & 29.0 & 18.5 & 11.5 & 21.0 & 35.5 & 43.5 & 32.6 \\
Llama-3.2-11B-Vision & 71.5 & 27.0 & 18.5 & 13.5 & 17.0 & 42.0 & 57.0 & 35.2 \\
Qwen3.5-397B         & 88.5 & 37.5 & 36.5 & 29.0 & 40.0 & 69.0 & 77.5 & 54.0 \\
Gemma-4-31B          & 86.5 & 17.5 & 22.0 & 18.0 & 22.5 & 32.5 & 47.0 & 35.1 \\
\bottomrule
\end{tabular}%
}
\end{table}

\textbf{Aspect Coverage and Engagement.} We tag each critique for seven
aesthetic aspects (composition, lighting, color, exposure, technique, mood,
and subject) with a keyword tagger: per aspect a hand-written list of
$15$--$25$ case-insensitive regular expressions, an aspect counting as
covered if any of them matches, applied identically to model and human
critiques. Appendix~\ref{sec:app-tagger} lists the terms and validates the
tagger against an independent LLM labeling of the same critiques, which
agrees with it on $79.6\%$ of decisions but detects more in the shorter, less
technical human critiques; the gaps below are therefore upper bounds. The headline contrast is
selectivity: humans mention composition in $48\%$ of critiques and most
other aspects far less often, addressing mood and exposure in barely $11\%$.
Models behave differently. When a single critique is requested they cover
nearly everything, and this breadth is a default rather than a reaction to
the prompt: with the aspect list removed under \textsc{Capped-Plain},
composition coverage stays above $97\%$ for all nine models, including
Kimi-K2.6 and GPT-5.5, and most aspects remain two to six times more
frequent than in human critiques (Table~\ref{tab:aspect-coverage}). The
aspect hint thus amplifies a disposition that is already present rather than
creating it. Nor is the breadth a byproduct of the word budget: cut to the
human range ($30$--$60$ words, third block), the models lower their coverage
without becoming selective, still mentioning most aspects well above the
human rate.

Asking for three critiques in a single pass is the only format that restores
a measure of selectivity, spreading the aspects more thinly across the three,
though coverage stays above the human rate on nearly every aspect
(Table~\ref{tab:aspect-coverage-multi} in the Appendix); it is also the only
format that reduces repetition across the three critiques. The full diversity analysis is
given in Appendix~\ref{sec:app-aspect-diversity}.
Comprehensiveness is therefore a stable default: the models bring exhaustive
coverage to any single critique they write and shed it only when the prompt
asks for several at once.


Coverage records whether an aspect is mentioned, not how substantively. To
probe depth, we have GPT-4o score each critique per aspect on a $0$--$3$
scale, from $0$ (not mentioned) to $3$ (specific observation plus actionable
insight) (Table~\ref{tab:aspect-engagement-judge}). On this measure the
judge rates model critiques above humans on every aspect: mean composition
depth is $1.56$ for humans versus $1.81$--$2.66$ for the models, and the
share of posts reaching substantive engagement (depth~$\geq 2$) on
\emph{subject} rises from $17.5\%$ for humans to $33$--$77.5\%$ for the
models. We report this as the judge's assessment rather than as ground
truth: whether such breadth and depth count as virtues depends on the judge
(Table~\ref{tab:usefulness-pref}), so we read these scores as a measure of
how uniformly and extensively the models engage each aspect, not as a
verdict on quality.



\section{Conclusion}
We tested how closely MLLM critiques of photographs match human ones, assessing
eight open-weight MLLMs and GPT-5.5 against ranked human references from
\texttt{r/photocritique} across eight prompt conditions, grounding controls,
and four LLM judges. Judged by reference-based similarity the picture is
deceptive: the metrics that track wording and meaning align only weakly with
human critiques, and the one favorable in absolute terms, a coarse
sentence-embedding cosine, registers little beyond topical overlap that our
controls tie to length and to the post text rather than to anything the model
saw. Four LLM judges and two annotators agree that model and human critiques
rarely say the same thing. The behavioral
picture exposes what the scores hide: held to a word limit the models still
write far more than people, touch nearly every aspect humans raise selectively,
and recycle content across critiques of one photo where humans offer distinct
ones.

\clearpage

{
    \small
    \bibliographystyle{ieeenat_fullname}
    \bibliography{main}
}

\onecolumn          
\clearpage
\appendix

\begin{center}
    {\Large\bfseries Supplementary Material:\\Can MLLMs Critique Like Humans? Evaluating Open-Ended Aesthetic Reasoning in Multimodal Large Language Models\par}
    \vspace{7em}
\end{center}

\renewcommand{\thefigure}{A\arabic{figure}}
\renewcommand{\thetable}{A\arabic{table}}
\setcounter{figure}{0}
\setcounter{table}{0}

\begin{promptbox}{Prompt Template --- \textsc{Open-Hinted}}
You are an experienced photographer browsing r/photocritique, an online community
where photographers share their work and receive constructive feedback.

\smallskip
A user posted this photo with the title: ``\pv{title}'' \pv{body}

\smallskip
Write one well-reasoned critique of this photo. Cover relevant aspects such as
composition, lighting, color, focus, or emotional impact. Be specific and constructive.
\end{promptbox}
\captionof{figure}{The \textsc{Open-Hinted} condition: persona framing and an aspect
hint, with no length constraint.}
\label{fig:prompt-open-hinted}

\bigskip
\begin{promptbox}{Prompt Template --- \textsc{Capped-Hinted}}
You are an experienced photographer browsing r/photocritique, an online community
where photographers share their work and receive constructive feedback.

\smallskip
A user posted this photo with the title: ``\pv{title}'' \pv{body}

\smallskip
Write one well-reasoned critique of this photo in 50--150 words. Cover relevant aspects
such as composition, lighting, color, focus, or emotional impact. Be specific and constructive.
\end{promptbox}
\captionof{figure}{The \textsc{Capped-Hinted} condition: persona framing and an aspect
hint with an added 50--150 word length constraint.}
\label{fig:prompt-capped-hinted}

\bigskip
\begin{promptbox}{Prompt Template --- \textsc{Multi-Hinted}}
You are an experienced photographer browsing r/photocritique, an online community
where photographers share their work and receive constructive feedback.

\smallskip
A user posted this photo with the title: ``\pv{title}'' \pv{body}

\smallskip
Write three independent critiques of this photo, each in 50--150 words. Focus each on a
different angle (e.g.\ one on composition, one on lighting and color, one on emotional
impact or storytelling). Be specific and constructive.

\smallskip
Format your response exactly as:\\
\texttt{Critique 1:} [critique]\quad \texttt{Critique 2:} [critique]\quad \texttt{Critique 3:} [critique]
\end{promptbox}
\captionof{figure}{The \textsc{Multi-Hinted} condition: a single pass producing three
critiques with explicit angle assignment.}
\label{fig:prompt-multi-hinted}

\bigskip
\begin{promptbox}{Prompt Template --- \textsc{Multi-Plain}}
A user posted this photo on r/photocritique with the title: ``\pv{title}'' \pv{body}

\smallskip
Write three different critiques of this photo as comments on this post. Each critique
should be 50--150 words.

\smallskip
Format your response exactly as:\\
\texttt{Critique 1:} [critique]\quad \texttt{Critique 2:} [critique]\quad \texttt{Critique 3:} [critique]
\end{promptbox}
\captionof{figure}{The \textsc{Multi-Plain} condition: a single pass producing three
critiques with no persona and no aspect hint.}
\label{fig:prompt-multi-plain}

\bigskip
\begin{promptbox}{Prompt Template --- \textsc{Unguided}}
A user posted this photo on r/photocritique with the title: ``\pv{title}'' \pv{body}

\smallskip
Write a critique of this photo.
\end{promptbox}
\captionof{figure}{The \textsc{Unguided} condition: minimal instruction with no persona,
no aspect hint, and no length constraint.}
\label{fig:prompt-unguided}

\begin{figure}[h]
\centering
\begin{promptbox}{Judge Prompt --- \textsc{Aspect-Engagement}}
You are evaluating a photography critique posted on r/photocritique.

A user posted this image with the title: ``\pv{title}'' \pv{body}

The critique below is one comment on this post.

\smallskip
CRITIQUE:\\
\pv{critique}
\smallskip

For each of the seven aspects below, rate how substantively the critique
engages with that aspect using this scale:

\smallskip
0 = Not mentioned at all\\
1 = Surface mention only (names the aspect without specifics, e.g., ``good
composition'' or ``nice colors'')\\
2 = Specific observation about this image (references visible elements:
e.g., ``the leading line of the railing draws the eye toward the subject'')\\
3 = Specific observation plus actionable insight or detailed analysis
(e.g., ``the leading line works but the subject is placed too centrally;
moving it to the right intersection would strengthen the composition'')
\smallskip

Aspects to score:
\begin{itemize}\itemsep0pt
  \item Composition (framing, balance, placement, leading lines)
  \item Lighting (light direction, quality, shadows, highlights)
  \item Color (color choices, palette, saturation, color relationships)
  \item Exposure (brightness, dynamic range, clipping)
  \item Technique (focus, sharpness, depth of field, camera settings)
  \item Mood (emotional impact, atmosphere, story)
  \item Subject (the main subject, framing of subject, subject treatment)
\end{itemize}

Respond in this EXACT format with no other text:\\
COMPOSITION: [0/1/2/3]\\
LIGHTING: [0/1/2/3]\\
COLOR: [0/1/2/3] \\
EXPOSURE: [0/1/2/3] \\
TECHNIQUE: [0/1/2/3] \\
MOOD: [0/1/2/3] \\
SUBJECT: [0/1/2/3] 
\end{promptbox}
\caption{Judge prompt for aspect engagement scoring. GPT-4o receives the image
alongside the text; \pv{title} and \pv{body} expand to the post title and body,
and \pv{critique} expands to the critique being evaluated. Scores are independent
per aspect on a 0--3 depth scale.}
\label{fig:judge-prompt-aspect}
\end{figure}

\tcbset{
  promptbox/.style={
    title after break={},
  }
}

\begin{figure}[t]
\centering
\begin{promptbox}{Judge Prompt --- \textsc{Usefulness}}
You are evaluating photography critiques on r/photocritique.

A user posted this image with the title: ``\pv{title}'' \pv{body}

\smallskip
Two critiques were written for this post:

\smallskip
Critique A: \pv{critique\_a}

\smallskip
Critique B: \pv{critique\_b}
\smallskip

Which critique would be more useful to the photographer? Consider:
\begin{itemize}\itemsep0pt
  \item Specificity to THIS image (vs. generic advice)
  \item Whether it offers actionable feedback
  \item Whether it engages with what's actually visible in the image
\end{itemize}

Respond in this EXACT format with no other text:\\
CHOICE: [A/B/Tie]
\end{promptbox}
\caption{Judge prompt for pairwise usefulness comparison. GPT-4o receives the
image alongside the text; \pv{title} and \pv{body} expand to the post title and
body, and \pv{critique\_a}, \pv{critique\_b} expand to the two critiques being
compared. Critique source (MLLM vs.\ human) is randomly assigned to position A
or B per pair to avoid order bias.}
\label{fig:judge-prompt-usefulness}
\end{figure}

\begin{figure}[t]
\centering
\begin{promptbox}{Judge Prompt --- \textsc{Similarity}}
You are comparing two photography critiques of the same photo, posted on
r/photocritique.

A user posted this image with the title: ``\pv{title}'' \pv{body}

Below are two critiques of this photo, labeled A and B.

\smallskip
CRITIQUE A:\\
\pv{critique\_a}

\smallskip
CRITIQUE B:\\
\pv{critique\_b}
\smallskip

Rate how similar Critique A and Critique B are in substance on a 1--5 scale.
Focus on whether they discuss the same things about the photo --- same
observations, same conclusions, same suggestions. Ignore differences in length,
style, vocabulary, and formatting.

\smallskip
1 = Completely different. They discuss entirely different aspects, reach
different conclusions, or would lead the photographer to make different
changes.\\
2 = Mostly different. Some shared observations but most of the content
differs.\\
3 = Half similar. Roughly equal mix of shared and differing observations.\\
4 = Mostly similar. Most observations and conclusions match; differ on a few
minor points.\\
5 = Nearly identical in substance. Same observations, same conclusions, same
suggestions --- even if phrased very differently.
\smallskip

Respond in this EXACT format with no other text:\\
SCORE: [1--5] -- [one sentence explanation]
\end{promptbox}
\caption{Judge prompt for pairwise substance similarity. GPT-4o receives the
image alongside the text; \pv{title} and \pv{body} expand to the post title
and body, and \pv{critique\_a}, \pv{critique\_b} expand to the two critiques
being compared. Critique source (MLLM vs.\ human) is randomly assigned to
position A or B per pair, with the same assignment used for both the GPT-4o
judge and human annotators on the matched subset.}
\label{fig:judge-prompt-similarity}
\end{figure}

\clearpage

\begin{table}[t]
\centering
\caption{Pairwise usefulness judgments from each of the four LLM judges under two output-length budgets ($100$ posts per cell, identical \textsc{Capped-Plain} prompt, differing only in the requested word range): share (\%) of pairs in which the judge preferred the MLLM critique, the human critique, or neither. The final block averages the four judges.}
\label{tab:usefulness-judge}
\resizebox{0.48\textwidth}{!}{%
\small
\begin{tabular}{lcccccc}
\toprule
& \multicolumn{3}{c}{Capped-Plain (50--150)} & \multicolumn{3}{c}{Capped-Plain (30--60)} \\
\cmidrule(lr){2-4} \cmidrule(lr){5-7}
Model & MLLM & Human & Tie & MLLM & Human & Tie \\
\midrule
\multicolumn{7}{l}{\textbf{Judge: \jicon{openai}\,GPT-4o}} \\
Qwen2-VL-7B      & 73.0 & 23.0 &  4.0 & 57.0 & 36.0 &  7.0 \\
InternVL3-8B     & 81.0 & 14.0 &  5.0 & 57.0 & 34.0 &  9.0 \\
Qwen3.5-397B     & 95.0 &  3.0 &  2.0 & 68.0 & 22.0 & 10.0 \\
Gemma-4-31B      & 91.0 &  0.0 &  9.0 & 74.0 & 16.0 & 10.0 \\
Kimi-K2.6        & 96.0 &  1.0 &  3.0 & 84.0 &  5.0 & 11.0 \\
GPT-5.5          & 96.0 &  1.0 &  3.0 & 86.0 &  6.0 &  8.0 \\
\midrule
\multicolumn{7}{l}{\textbf{Judge: \jicon{openai}\,GPT-5.5}} \\
Qwen2-VL-7B      & 39.0 & 61.0 &  0.0 & 29.0 & 71.0 &  0.0 \\
InternVL3-8B     & 38.0 & 62.0 &  0.0 & 30.0 & 70.0 &  0.0 \\
Qwen3.5-397B     & 78.0 & 22.0 &  0.0 & 63.0 & 37.0 &  0.0 \\
Gemma-4-31B      & 84.0 & 16.0 &  0.0 & 63.0 & 37.0 &  0.0 \\
Kimi-K2.6        & 95.0 &  5.0 &  0.0 & 87.0 & 13.0 &  0.0 \\
GPT-5.5          & 99.0 &  1.0 &  0.0 & 91.0 &  9.0 &  0.0 \\
\midrule
\multicolumn{7}{l}{\textbf{Judge: \jicon{googlegemini}\,Gemini-3.1-Pro}} \\
Qwen2-VL-7B      & 11.0 & 88.0 &  1.0 & 15.0 & 85.0 &  0.0 \\
InternVL3-8B     & 12.0 & 88.0 &  0.0 & 16.0 & 84.0 &  0.0 \\
Qwen3.5-397B     & 73.0 & 27.0 &  0.0 & 57.0 & 43.0 &  0.0 \\
Gemma-4-31B      & 79.0 & 20.0 &  1.0 & 60.0 & 40.0 &  0.0 \\
Kimi-K2.6        & 94.0 &  6.0 &  0.0 & 81.0 & 19.0 &  0.0 \\
GPT-5.5          & 97.0 &  3.0 &  0.0 & 93.0 &  6.0 &  1.0 \\
\midrule
\multicolumn{7}{l}{\textbf{Judge: \jicon{claude}\,Claude-Opus-5}} \\
Qwen2-VL-7B      & 11.0 & 89.0 &  0.0 & 14.0 & 86.0 &  0.0 \\
InternVL3-8B     &  9.0 & 91.0 &  0.0 &  8.0 & 92.0 &  0.0 \\
Qwen3.5-397B     & 52.0 & 48.0 &  0.0 & 42.0 & 58.0 &  0.0 \\
Gemma-4-31B      & 67.0 & 33.0 &  0.0 & 46.0 & 54.0 &  0.0 \\
Kimi-K2.6        & 84.0 & 16.0 &  0.0 & 64.0 & 36.0 &  0.0 \\
GPT-5.5          & 85.0 & 15.0 &  0.0 & 75.0 & 25.0 &  0.0 \\
\midrule
\multicolumn{7}{l}{\textbf{Mean across the four judges}} \\
\textit{Qwen2-VL-7B}     & 33.5 & 65.2 &  1.2 & 28.7 & 69.5 &  1.8 \\
\textit{InternVL3-8B}    & 35.0 & 63.8 &  1.2 & 27.8 & 70.0 &  2.2 \\
\textit{Qwen3.5-397B}    & 74.5 & 25.0 &  0.5 & 57.5 & 40.0 &  2.5 \\
\textit{Gemma-4-31B}     & 80.2 & 17.2 &  2.5 & 60.8 & 36.8 &  2.5 \\
\textit{Kimi-K2.6}       & 92.2 &  7.0 &  0.8 & 79.0 & 18.2 &  2.8 \\
\textit{GPT-5.5}         & 94.2 &  5.0 &  0.8 & 86.2 & 11.5 &  2.2 \\
\bottomrule
\end{tabular}%
}
\end{table}

\begin{table}[t]
\centering
\caption{Per-judge similarity ratings on the same $100$ posts: mean rating of how similar the MLLM critique and the matched human critique are \emph{in substance} on a $1$--$5$ scale ($1$ completely different, $2$ mostly different, $5$ essentially the same content), ignoring length, style and vocabulary. \textit{Mean} averages the four judges; $\geq 4$ is the share of pairs rated $4$ or $5$.}
\label{tab:sim-judge}
\resizebox{0.7\textwidth}{!}{%
\small
\begin{tabular}{lccccrr}
\toprule
Model & \jicon{openai}\,GPT-4o & \jicon{openai}\,GPT-5.5 & \jicon{googlegemini}\,Gemini-3.1-Pro & \jicon{claude}\,Claude-Opus-5 & \textit{Mean} & $\geq 4$ (\%) \\
\midrule
Qwen2-VL-7B      & 2.42 & 2.19 & 1.73 & 1.92 & 2.06 & 5.0 \\
InternVL3-8B     & 2.45 & 2.21 & 1.71 & 1.94 & 2.08 & 3.0 \\
Qwen3.5-397B     & 2.72 & 2.54 & 2.30 & 2.40 & 2.49 & 17.5 \\
Gemma-4-31B      & 2.81 & 2.52 & 2.18 & 2.34 & 2.46 & 15.8 \\
Kimi-K2.6        & 2.77 & 2.54 & 2.29 & 2.41 & 2.50 & 18.8 \\
GPT-5.5          & 2.97 & 2.56 & 2.48 & 2.58 & 2.65 & 19.0 \\
\bottomrule
\end{tabular}%
}
\end{table}

\begin{table}[t]
\centering
\caption{Agreement with the human annotators on the $150$ annotated pairs of
Table~\ref{tab:human-validation}. The first row compares the two annotators
with each other; each judge row compares the judge's ratings against each
annotator in turn, averaged over the two. \textsc{Exact} and $\pm1$ are
agreement rates, $\rho$ is Spearman's.}
\label{tab:judge-agreement}
\resizebox{0.4\textwidth}{!}{%
\small
\begin{tabular}{lccc}
\toprule
Rater pair & Exact (\%) & $\pm1$ (\%) & $\rho$ \\
\midrule
A1 vs.\ A2     & 31.8 & 73.6 & 0.43 \\
\midrule
\jicon{openai}\,GPT-4o & 35.2 & 85.9 & 0.53 \\
\jicon{openai}\,GPT-5.5 & 31.2 & 85.9 & 0.57 \\
\jicon{googlegemini}\,Gemini-3.1-Pro & 30.9 & 73.0 & 0.58 \\
\jicon{claude}\,Claude-Opus-5 & 28.9 & 79.2 & 0.54 \\
\bottomrule
\end{tabular}%
}
\end{table}

\section{Full LLM-as-Judge Results}
\label{sec:app-judges}

Table~\ref{tab:usefulness-judge} expands the usefulness study of
Section~\ref{sec:results} to the full per-judge record, adding the human and
tie shares under both output-length budgets. Each judge saw the image, the
post title and body, and an anonymized pair of critiques, one MLLM generated
and one human, in randomized order, and selected which would be more useful
to the photographer (prompt in Fig.~\ref{fig:judge-prompt-usefulness}).
Shortening the model output from $50$--$150$ to $30$--$60$ words lowers the
MLLM preference for every judge, by $18.8$ points for GPT-4o, $12.4$ for
GPT-5.5, $6.2$ for Gemini-3.1-Pro and $7.8$ for Claude-Opus-5, and raises
four-way agreement on individual pairs from $50.0\%$ to $55.2\%$.

\vspace{0.2cm}
Table~\ref{tab:sim-judge} reports the similarity counterpart on the same
$100$ posts: each judge rated how similar the MLLM critique and the matched
human critique are in substance, using the rubric of
Fig.~\ref{fig:judge-prompt-similarity}. Table~\ref{tab:judge-agreement}
completes the picture with rater agreement on the $150$ pairs annotated by
humans in Table~\ref{tab:human-validation}. Gemini-3.1-Pro declined $10$ of
the $150$ annotated pairs on safety grounds and is scored on the remaining
$140$.

\begin{table}[t]
\centering
\caption{Image-grounding test under BERTScore ($N{=}500$ posts, identical post set for every row). Columns are as in Table~\ref{tab:grounding}: critiques written with the correct image, a shuffled image and no image, scored against the same human references; the largest of the three is bold. Kimi-K2.6 and GPT-5.5 were run in the primary condition only.}
\label{tab:grounding-bert}
\resizebox{0.55\textwidth}{!}{%
\small
\begin{tabular}{lccccc}
\toprule
Model & C-vs-H & S-vs-H & T-vs-H & $\Delta_{\mathrm{S}}$ & $\Delta_{\mathrm{T}}$ \\
\midrule
\multicolumn{6}{l}{\textit{Open-Hinted}} \\
\midrule
Qwen2-VL-7B          & $\mathbf{-0.055}$ & $-0.075$ & $-0.059$ & $+0.019$ & $+0.004$ \\
LLaVA-1.6-Mistral-7B & $-0.047$ & $-0.059$ & $\mathbf{-0.009}$ & $+0.011$ & $-0.038$ \\
LLaVA-OneVision-7B   & $\mathbf{-0.043}$ & $-0.060$ & $-0.069$ & $+0.017$ & $+0.026$ \\
InternVL3-8B         & $\mathbf{-0.064}$ & $-0.080$ & $-0.085$ & $+0.015$ & $+0.021$ \\
Llama-3.2-11B-Vision & $-0.036$ & $\mathbf{-0.034}$ & $-0.046$ & $-0.002$ & $+0.011$ \\
Qwen3.5-397B         & $\mathbf{-0.051}$ & $-0.076$ & $-0.076$ & $+0.025$ & $+0.025$ \\
Gemma-4-31B          & $\mathbf{-0.027}$ & $-0.048$ & $-0.077$ & $+0.021$ & $+0.050$ \\
\midrule
\multicolumn{6}{l}{\textit{Capped-Hinted}} \\
\midrule
Qwen2-VL-7B          & $\mathbf{0.074}$ & $0.053$ & $0.070$ & $+0.022$ & $+0.005$ \\
LLaVA-1.6-Mistral-7B & $0.040$ & $0.026$ & $\mathbf{0.077}$ & $+0.014$ & $-0.036$ \\
LLaVA-OneVision-7B   & $\mathbf{0.040}$ & $0.022$ & $0.031$ & $+0.018$ & $+0.009$ \\
InternVL3-8B         & $\mathbf{0.043}$ & $0.024$ & $0.014$ & $+0.019$ & $+0.029$ \\
Llama-3.2-11B-Vision & $0.070$ & $0.059$ & $\mathbf{0.088}$ & $+0.011$ & $-0.018$ \\
Qwen3.5-397B         & $0.068$ & $0.041$ & $\mathbf{0.070}$ & $+0.027$ & $-0.001$ \\
Gemma-4-31B          & $\mathbf{0.105}$ & $0.079$ & $0.047$ & $+0.026$ & $+0.058$ \\
\midrule
\multicolumn{6}{l}{\textit{Capped-Plain}} \\
\midrule
Qwen2-VL-7B          & $\mathbf{0.100}$ & $0.075$ & $0.097$ & $+0.024$ & $+0.003$ \\
LLaVA-1.6-Mistral-7B & $0.056$ & $0.043$ & $\mathbf{0.090}$ & $+0.013$ & $-0.034$ \\
LLaVA-OneVision-7B   & $\mathbf{0.065}$ & $0.043$ & $0.046$ & $+0.023$ & $+0.019$ \\
InternVL3-8B         & $\mathbf{0.073}$ & $0.052$ & $0.040$ & $+0.021$ & $+0.033$ \\
Llama-3.2-11B-Vision & $\mathbf{0.084}$ & $0.068$ & $0.081$ & $+0.016$ & $+0.003$ \\
Qwen3.5-397B         & $\mathbf{0.081}$ & $0.052$ & $0.072$ & $+0.029$ & $+0.009$ \\
Gemma-4-31B          & $\mathbf{0.103}$ & $0.079$ & $0.037$ & $+0.025$ & $+0.066$ \\
Kimi-K2.6            & $\mathbf{0.085}$ & $0.059$ & $0.036$ & $+0.026$ & $+0.049$ \\
GPT-5.5              & $\mathbf{0.099}$ & $0.070$ & $0.062$ & $+0.028$ & $+0.036$ \\
\bottomrule
\end{tabular}%
}
\end{table}



\section{Image Grounding under BERTScore}
\label{sec:grounding-bertscore}
Table~\ref{tab:grounding-bert} repeats the image-grounding test of
Section~\ref{sec:results} using BERTScore in place of SBERT-cos. The
qualitative pattern replicates under the stricter metric. Correct-image
critiques score higher than shuffled-image ones in $22$ of $23$ cells
($\Delta_{\mathrm{S}}$ mean $+0.019$), confirming that grounding is real,
and $\Delta_{\mathrm{T}}$ again tracks model capability, from $-0.038$ for
LLaVA-1.6-Mistral-7B up to $+0.066$ for Gemma-4-31B, with the frontier
models at the image-dependent end. Two differences from the cosine results
are worth noting. First, the grounding gap is much smaller in absolute
terms, around $0.02$ rather than the $0.10$ seen under SBERT-cos, so the
benefit of seeing the correct image is real but slight. Second, under
\textsc{Open-Hinted} the long, uncapped critiques fall below the unrelated-text level against the human references (negative C-vs-H, S-vs-H and T-vs-H) regardless of what
the model saw, whereas the length-capped conditions are weakly positive. How
much a model depends on the photograph is therefore a property of the model
rather than of the metric or the task.



\begin{table}[t]
\centering
\caption{Aspect coverage (\%) under the multi-critique conditions ($N{=}1{,}227$). Each cell is the fraction of critiques that address the given aspect. \textit{Human} is the reference baseline. \textsc{Multi-Plain} and \textsc{Multi-Hinted} both produce three critiques in a single pass, without and with an explicit aspect hint respectively. \textit{Avg.} is the mean across the seven aspects.}
\label{tab:aspect-coverage-multi}
\resizebox{0.55\textwidth}{!}{%
\begin{tabular}{lcccccccc}
\toprule
& Comp. & Light. & Color & Expo. & Tech. & Mood & Subj. & \textit{Avg.} \\
\midrule
\textit{Human}       & 48.2 & 23.0 & 22.1 & 11.0 & 33.7 & 10.5 & 35.8 & 26.3 \\
\midrule
\multicolumn{9}{l}{\textit{Multi-Plain}} \\
\midrule
Qwen2-VL-7B          & 90.5 & 60.4 & 51.7 & 30.9 & 41.8 & 45.9 & 66.1 & 55.3 \\
LLaVA-1.6-Mistral-7B & 94.7 & 61.6 & 56.9 & 23.6 & 55.5 & 67.2 & 72.5 & 61.7 \\
LLaVA-OneVision-7B   & 88.7 & 50.4 & 48.9 & 27.3 & 52.3 & 51.9 & 72.5 & 56.0 \\
InternVL3-8B         & 88.3 & 61.9 & 49.7 & 25.8 & 52.2 & 52.8 & 61.1 & 56.0 \\
Llama-3.2-11B-Vision & 90.1 & 46.1 & 45.9 & 19.8 & 53.4 & 60.0 & 59.2 & 53.5 \\
Qwen3.5-397B         & 95.4 & 66.6 & 69.6 & 31.5 & 69.3 & 75.3 & 70.1 & 68.3 \\
Gemma-4-31B          & 94.6 & 58.0 & 50.0 & 26.1 & 62.2 & 55.6 & 65.8 & 58.9 \\
\midrule
\multicolumn{9}{l}{\textit{Multi-Hinted}} \\
\midrule
Qwen2-VL-7B          & 73.2 & 49.6 & 46.8 & 19.1 & 22.6 & 60.5 & 68.5 & 48.6 \\
LLaVA-1.6-Mistral-7B & 77.9 & 52.6 & 49.7 & 17.2 & 36.5 & 75.4 & 71.6 & 54.4 \\
LLaVA-OneVision-7B   & 72.9 & 47.3 & 41.3 & 20.8 & 34.2 & 60.3 & 70.8 & 49.7 \\
InternVL3-8B         & 66.2 & 46.9 & 40.3 & 14.2 & 30.9 & 63.4 & 67.2 & 47.0 \\
Llama-3.2-11B-Vision & 86.0 & 39.1 & 38.9 & 9.1 & 28.4 & 70.7 & 64.2 & 48.1 \\
Qwen3.5-397B         & 74.0 & 50.6 & 46.1 & 18.0 & 48.8 & 75.5 & 70.1 & 54.7 \\
Gemma-4-31B          & 84.2 & 45.0 & 37.7 & 19.0 & 47.7 & 64.3 & 70.0 & 52.6 \\
\bottomrule
\end{tabular}%
}
\end{table}

\begin{table*}[t]
\centering
\caption{Mean pairwise self-similarity among the three critiques generated for the same post ($N{=}1{,}227$); higher means less diverse. The first four conditions compare three independently sampled critiques, the next two the three critiques produced in a single pass, and the last two the length-matched variants of \textsc{Capped-Plain}. \textit{Human} is the mean pairwise similarity among the three human critiques per post. Llama-3.2-11B-Vision uses a repetition-penalty variant under \textsc{Multi-Plain} and \textsc{Multi-Hinted}, where the base model degenerates into repeating itself.}
\label{tab:diversity}
\setlength{\tabcolsep}{2pt}
\resizebox{0.95\textwidth}{!}{%
\small
\begin{tabular}{l cccccccc c cccccccc}
\toprule
& \multicolumn{8}{c}{ROUGE-L} && \multicolumn{8}{c}{SBERT-cos} \\
\cmidrule(lr){2-9} \cmidrule(lr){11-18}
Model & O-Hint & C-Hint & C-Plain & U-Guid & M-Plain & M-Hint & C-P(30--60) & C-P(20--45) && O-Hint & C-Hint & C-Plain & U-Guid & M-Plain & M-Hint & C-P(30--60) & C-P(20--45) \\
\midrule
Qwen2-VL-7B          & 0.616 & 0.777 & 0.790 & 0.633 & 0.245 & 0.237 & 0.856 & 0.867 && 0.967 & 0.965 & 0.963 & 0.966 & 0.745 & 0.659 & 0.967 & 0.966 \\
LLaVA-1.6-Mistral-7B & 0.295 & 0.302 & 0.286 & 0.263 & 0.241 & 0.230 & 0.308 & 0.325 && 0.850 & 0.797 & 0.799 & 0.843 & 0.791 & 0.688 & 0.807 & 0.810 \\
LLaVA-OneVision-7B   & 0.270 & 0.283 & 0.278 & 0.247 & 0.198 & 0.191 & 0.323 & 0.327 && 0.872 & 0.844 & 0.829 & 0.862 & 0.712 & 0.617 & 0.817 & 0.793 \\
InternVL3-8B         & 0.270 & 0.307 & 0.314 & 0.264 & 0.161 & 0.145 & 0.413 & 0.422 && 0.909 & 0.901 & 0.885 & 0.904 & 0.715 & 0.597 & 0.872 & 0.838 \\
Llama-3.2-11B-Vision & 0.297 & 0.323 & 0.310 & 0.307 & 0.221 & 0.216 & 0.302 & 0.309 && 0.854 & 0.843 & 0.830 & 0.838 & 0.761 & 0.735 & 0.755 & 0.732 \\
Qwen3.5-397B         & 0.286 & 0.356 & 0.362 & 0.307 & 0.171 & 0.161 & 0.421 & 0.423 && 0.919 & 0.898 & 0.883 & 0.930 & 0.778 & 0.644 & 0.829 & 0.812 \\
Gemma-4-31B          & 0.381 & 0.467 & 0.474 & 0.372 & 0.211 & 0.199 & 0.544 & 0.577 && 0.917 & 0.893 & 0.888 & 0.920 & 0.739 & 0.623 & 0.849 & 0.843 \\
\midrule
\textit{Human} & \multicolumn{8}{c}{0.120} && \multicolumn{8}{c}{0.436} \\
\bottomrule
\end{tabular}
}
\end{table*}

\section{Multi-Critique Coverage and Full Diversity Results}
\label{sec:app-aspect-diversity}

Table~\ref{tab:aspect-coverage-multi} extends the aspect-coverage analysis
of Section~\ref{sec:results} to the two multi-critique conditions, in which
a model produces three critiques in a single pass. Coverage falls on every
aspect relative to the single-critique conditions of
Table~\ref{tab:aspect-coverage}, and \textsc{Multi-Hinted} compresses it
further, since assigning each critique a distinct angle pulls it toward one
or two aspects. Even so, the mean coverage across the seven aspects remains
roughly twice the human level for every model ($47.0$--$68.3$ against
$26.3$).

\vspace{0.2cm}
Table~\ref{tab:diversity} reports the full within-post self-similarity
results for the seven open-weight models under all eight conditions, on
both the lexical and the embedding metric. Three independently sampled
critiques of the same photo are near-duplicates in meaning (SBERT-cos
$0.797$--$0.967$ against a human $0.436$), the single-pass conditions
roughly halve the redundancy ($0.597$--$0.791$) without reaching the human
level, and cutting the word budget to the human range raises lexical
self-similarity for six of the seven models, so the length-matched
conditions buy shorter critiques at the price of even less variety.

\begin{table*}[t]
\centering
\caption{Effect of the aggregation rule under the primary \textsc{Capped-Plain}
condition ($N{=}1{,}227$). \emph{Row} is the row-wise maximum used throughout the
paper. \emph{Eq.2} scores each model critique against two references rather than
three, averaged over the three subsets, matching the number available to a
leave-one-out human candidate; the human value is unchanged by this column for
that reason. \emph{1:1} assigns the three model critiques to three distinct
references by optimal assignment. Both stricter protocols lower every score, and
neither moves any model across the human agreement baseline.}
\label{tab:aggregation-robustness}
\setlength{\tabcolsep}{4pt}
\resizebox{0.9\textwidth}{!}{%
\small
\begin{tabular}{lccccccccccccc}
\toprule
& \multicolumn{3}{c}{ROUGE-L} & \multicolumn{3}{c}{BERTScore}
& \multicolumn{3}{c}{SBERT-cos} & \multicolumn{3}{c}{BLEURT} \\
\cmidrule(lr){2-4} \cmidrule(lr){5-7} \cmidrule(lr){8-10} \cmidrule(lr){11-13}
Model & Row & Eq.2 & 1:1 & Row & Eq.2 & 1:1 & Row & Eq.2 & 1:1 & Row & Eq.2 & 1:1 \\
\midrule
Qwen2-VL-7B            & 0.142 & 0.130 & 0.109 & $\phantom{-}0.099$ & $\phantom{-}0.080$ & $\phantom{-}0.040$ & 0.621 & 0.580 & 0.500 & $-0.60$ & $-0.66$ & $-0.80$ \\
LLaVA-1.6-Mistral-7B   & 0.134 & 0.122 & 0.105 & $\phantom{-}0.055$ & $\phantom{-}0.038$ & $\phantom{-}0.009$ & 0.594 & 0.555 & 0.496 & $-0.70$ & $-0.76$ & $-0.84$ \\
LLaVA-OneVision-7B     & 0.132 & 0.121 & 0.105 & $\phantom{-}0.066$ & $\phantom{-}0.048$ & $\phantom{-}0.018$ & 0.611 & 0.572 & 0.509 & $-0.63$ & $-0.69$ & $-0.80$ \\
InternVL3-8B           & 0.128 & 0.117 & 0.102 & $\phantom{-}0.074$ & $\phantom{-}0.056$ & $\phantom{-}0.024$ & 0.615 & 0.573 & 0.503 & $-0.60$ & $-0.67$ & $-0.78$ \\
Llama-3.2-11B-Vision   & 0.141 & 0.128 & 0.110 & $\phantom{-}0.083$ & $\phantom{-}0.065$ & $\phantom{-}0.034$ & 0.626 & 0.585 & 0.520 & $-0.62$ & $-0.68$ & $-0.77$ \\
Qwen3.5-397B-A17B      & 0.135 & 0.124 & 0.110 & $\phantom{-}0.082$ & $\phantom{-}0.064$ & $\phantom{-}0.033$ & 0.641 & 0.599 & 0.525 & $-0.85$ & $-0.92$ & $-1.03$ \\
Gemma-4-31B            & 0.147 & 0.135 & 0.118 & $\phantom{-}0.103$ & $\phantom{-}0.085$ & $\phantom{-}0.050$ & 0.648 & 0.605 & 0.530 & $-0.88$ & $-0.94$ & $-1.05$ \\
Kimi-K2.6              & 0.139 & 0.128 & 0.114 & $\phantom{-}0.085$ & $\phantom{-}0.067$ & $\phantom{-}0.036$ & 0.632 & 0.592 & 0.526 & $-0.93$ & $-0.99$ & $-1.10$ \\
GPT-5.5                & 0.143 & 0.132 & 0.116 & $\phantom{-}0.099$ & $\phantom{-}0.080$ & $\phantom{-}0.045$ & 0.649 & 0.607 & 0.532 & $-0.88$ & $-0.94$ & $-1.05$ \\
\midrule
\textit{Human agreement} & 0.147 & 0.147 & 0.120 & $\phantom{-}0.114$ & $\phantom{-}0.114$ & $\phantom{-}0.069$ & 0.514 & 0.514 & 0.436 & $-1.08$ & $-1.08$ & $-1.17$ \\
\bottomrule
\end{tabular}
}
\end{table*}

\begin{table}[t]
\centering
\caption{Number of models, out of nine, scoring above the human agreement
baseline under each aggregation rule (\textsc{Capped-Plain}, $N{=}1{,}227$). The
verdict is identical under every protocol for every metric.}
\label{tab:aggregation-verdict}
\resizebox{0.40\textwidth}{!}{%
\small
\begin{tabular}{lccc}
\toprule
Metric & Row-max & Equal 2 ref. & One-to-one \\
\midrule
ROUGE-L    & 0/9 & 0/9 & 0/9 \\
BERTScore  & 0/9 & 0/9 & 0/9 \\
SBERT-cos  & 9/9 & 9/9 & 9/9 \\
BLEURT     & 9/9 & 9/9 & 9/9 \\
\bottomrule
\end{tabular}%
}
\end{table}

\section{Robustness of the Max-Mapping Aggregation}
\label{sec:aggregation-robustness}

The max-mapping aggregation of Section~\ref{sec:setup} reads the model scores against a leave-one-out
human quantity in order to calibrate their range, and observes that the metrics
place the models at or above that quantity, a reading we argue is a measurement
artifact rather than evidence of human-level critique. The aggregation used to
produce those numbers is not perfectly symmetric between models and humans,
which raises the question of whether the inflation we attribute to the metrics
is instead produced by the matching rule. Two asymmetries exist. First, each
model critique takes a row-wise maximum over all three human references, whereas
a leave-one-out human candidate has only two available, because it consumes one
of its own. Second, row-wise maximization is many-to-one, so all three of a
model's critiques may select the same human reference. Both favor the model
side.

We therefore re-aggregate the stored pairwise similarity matrices under two
stricter protocols. No re-scoring is involved; each variant is a different
reduction of the same numbers.

\vspace{0.2cm}
\textbf{Equal reference counts.} Each model critique is scored against a
two-reference subset rather than all three, averaged over the three possible
subsets, so that both sides take a maximum over two references. The human
quantity is unchanged, since a leave-one-out candidate already has exactly two
references available. Equalizing the count is thus a one-sided correction
applied to the models.

\vspace{0.2cm}
\textbf{One-to-one matching.} The three model critiques are assigned to three
distinct human references by optimal assignment, so no reference may be claimed
twice. The corresponding human quantity is the best derangement among the three
human critiques, in which every critique is matched to a different one.

\vspace{0.2cm}
Table~\ref{tab:aggregation-robustness} reports both variants alongside the
published aggregation for all four metrics, and
Table~\ref{tab:aggregation-verdict} summarizes how many models fall on each side
of the human quantity under each protocol.

\vspace{0.2cm}
\textbf{The calibration reading does not depend on the aggregation.} Both
stricter protocols lower the absolute scores of every model, as expected when
references become scarcer or may not be reused. They do not move any model
across the human quantity. SBERT-cos and BLEURT place all nine models above it
under all three protocols, and ROUGE-L and BERTScore place all nine at or
below it under all three. Not one model changes side. The asymmetry therefore does not
explain the inflated scores: they come from the metrics, not from the matching
rule.

\vspace{0.2cm}
\textbf{One-to-one matching also quantifies redundancy.} Forbidding repeated
references costs the models far more than it costs the humans. Qwen2-VL-7B falls
from $0.0986$ to $0.0401$ on BERTScore, a loss of roughly $59\%$, while the
human quantity falls from $0.1139$ to $0.0694$, a loss of roughly $39\%$. The
models lose more because their three critiques repeatedly matched the same
human reference, which is what one expects of a model that produces three
near-identical critiques of the same photograph. The row-wise maximum therefore
rewards the redundancy documented in the within-image diversity analysis
(Table~\ref{tab:diversity}), and one-to-one matching makes the similarity metric
sensitive to it.

\vspace{0.2cm}
Neither variant is the uniquely correct protocol. A leave-one-out human
candidate must consume one of its own references, so no exactly symmetric
comparison exists: equalizing the reference count removes the counting
advantage, while one-to-one matching removes the reuse advantage. We report both
because the conclusions of this paper survive each of them. Following this
analysis we describe the leave-one-out quantity as a human \emph{agreement
baseline} rather than a ceiling, since models exceed it on two of the four
metrics and it is used here only to calibrate the range of the scores.

\begin{table}[t]
\centering
\caption{Precision of the human agreement baseline ($N{=}1{,}227$ posts).
\emph{Per-image SD} is the standard deviation of the per-post score and
describes how noisy a single photograph is; \emph{SE} and the two intervals
describe the precision of the reported mean. The bootstrap uses $10{,}000$
resamples over posts and agrees with the normal approximation to four decimal
places.}
\label{tab:ceiling-ci}
\resizebox{0.6\textwidth}{!}{%
\small
\begin{tabular}{lcccc}
\toprule
Metric & Mean & Per-image SD & SE & 95\% CI (bootstrap) \\
\midrule
ROUGE-L   & $\phantom{-}0.1472$ & 0.0426 & 0.0012 & $[\phantom{-}0.1449, \phantom{-}0.1495]$ \\
BERTScore & $\phantom{-}0.1139$ & 0.0830 & 0.0024 & $[\phantom{-}0.1093, \phantom{-}0.1185]$ \\
SBERT-cos & $\phantom{-}0.5142$ & 0.1307 & 0.0037 & $[\phantom{-}0.5069, \phantom{-}0.5216]$ \\
BLEURT    & $-1.0773$ & 0.1966 & 0.0056 & $[-1.0882, -1.0661]$ \\
\bottomrule
\end{tabular}%
}
\end{table}

\begin{table}[t]
\centering
\caption{Human agreement baseline as a function of the number of critiques per
post, on the $400$ posts with at least six qualifying non-OP comments. The post
set is held fixed across $K$, so the columns differ only in how many critiques
are used. $K{=}3$ is the value reported elsewhere in the paper. The baseline
rises monotonically with $K$ while the per-critique increment roughly halves at
each step, indicating a converging curve and a baseline that is understated at
$K{=}3$ by a bounded amount.}
\label{tab:ceiling-refcount}
\resizebox{0.6\textwidth}{!}{%
\small
\begin{tabular}{lccccc c}
\toprule
Metric & $K{=}2$ & $K{=}3$ & $K{=}4$ & $K{=}5$ & $K{=}6$ & $K{=}3 \rightarrow K{=}6$ \\
\midrule
ROUGE-L   & $\phantom{-}0.1205$ & $\phantom{-}0.1437$ & $\phantom{-}0.1562$ & $\phantom{-}0.1646$ & $\phantom{-}0.1717$ & $+0.0281$ \\
BERTScore & $\phantom{-}0.0804$ & $\phantom{-}0.1161$ & $\phantom{-}0.1350$ & $\phantom{-}0.1469$ & $\phantom{-}0.1553$ & $+0.0392$ \\
SBERT-cos & $\phantom{-}0.4509$ & $\phantom{-}0.5093$ & $\phantom{-}0.5400$ & $\phantom{-}0.5526$ & $\phantom{-}0.5620$ & $+0.0527$ \\
BLEURT    & $-1.1626$ & $-1.0592$ & $-1.0059$ & $-0.9801$ & $-0.9549$ & $+0.1044$ \\
\bottomrule
\end{tabular}%
}
\end{table}

\section{Stability of the Human Agreement Baseline}
\label{sec:ceiling-sensitivity}

The leave-one-out baseline is estimated from three critiques per post, which
raises two separate questions: how precisely it is estimated, and how it would
change with more critiques. The first is about variance, the second about bias.

\vspace{0.2cm}
\textbf{Per-image estimates are noisy; the reported mean is not.} On BERTScore
the per-post standard deviation is $0.083$ against a mean of $0.114$, and $8\%$
of posts fall below zero, meaning three human critiques of one photograph agree
less than unrelated text. This dispersion is not measurement error. It reflects
the property that motivates a multiple-reference protocol in the first place,
namely that competent critics of the same photograph often discuss entirely
different things. Every quantity we report is a mean over $1{,}227$ posts, so the
relevant uncertainty is that of the mean. Table~\ref{tab:ceiling-ci} gives the
standard error together with a $10{,}000$ resample bootstrap over posts. The two
intervals agree to four decimal places, and both are an order of magnitude
narrower than the spread of the model scores they are used to interpret: the
ROUGE-L interval spans less than $0.005$, while the model scores span
$0.128$ to $0.147$.

\vspace{0.2cm}
\textbf{Three critiques give a conservative baseline, and we can measure by how
much.} A confidence interval cannot answer the second question, since every
resample still uses the same three critiques. The benchmark retains only the top
three qualifying comments per post, but $400$ posts had at least six available.
On those posts we recompute the baseline using exactly $K$ critiques for
$K=2,\dots,6$, holding the post set fixed so the comparison is not confounded by
which posts attract many comments (Table~\ref{tab:ceiling-refcount}). The
baseline rises monotonically with $K$ on every metric while the increments fall
by roughly half at each step, so the curve is converging rather than diverging.
Moving from three to six critiques adds $0.028$ on ROUGE-L, $0.039$ on
BERTScore, $0.053$ on SBERT-cos and $0.104$ on BLEURT. The value at $K{=}3$ is
therefore understated by a measurable amount rather than unstable.

\vspace{0.2cm}
\textbf{The conclusions survive a debiased baseline.} We retain $K{=}3$ for two
reasons: it matches the number of references available to a model critique, so
the comparison stays symmetric in the sense discussed in
Section~\ref{sec:aggregation-robustness}, and a $K{=}6$ value can be computed
only on the third of posts that attracted at least six comments, which are more
discussed than the benchmark as a whole. Read as a correction rather than a
replacement, the curve favors our interpretation on both sides of the metric
split. On SBERT-cos every model scores between $0.594$ and $0.649$, above even
the $K{=}6$ baseline of $0.562$, so the observation that the metrics place models
at or above human agreement is not an artifact of having only three critiques.
On ROUGE-L and BERTScore, where models already fall at or below the baseline,
a higher baseline moves them further below: on the $400$-post subset the
ROUGE-L baseline rises from $0.144$ at $K{=}3$ to $0.156$ at $K{=}4$, already
clearly above the best model's $0.147$. A baseline estimated from more
critiques would make the models look relatively worse, not better.

\begin{table}[t]
\centering
\caption{Aspect coverage (\%) measured two ways on the $200$ judged posts
under \textsc{Multi-Plain}: keyword tagger / LLM judge (score $\geq 1$). The
two agree on $79.6\%$ of the $33{,}530$ critique--aspect decisions. Coverage
here is computed on the judged subset and so differs from
Table~\ref{tab:aspect-coverage}, which uses all $1{,}227$ posts under
\textsc{Capped-Plain}.}
\label{tab:tagger-validation}
\resizebox{0.9\textwidth}{!}{%
\small
\begin{tabular}{lccccccc}
\toprule
& Comp. & Light. & Color & Expo. & Tech. & Mood & Subj. \\
\midrule
\textit{Human}         & 53.1/68.9 & 24.7/29.9 & 22.5/31.9 & 10.8/27.2 & 35.0/44.2 & 12.8/49.1 & 35.8/64.4 \\
\midrule
Qwen2-VL-7B            & 91.3/84.0 & 64.0/62.0 & 51.3/55.8 & 30.0/40.3 & 43.7/46.8 & 43.8/81.5 & 67.5/87.0 \\
LLaVA-1.6-Mistral-7B   & 93.3/94.5 & 61.4/60.4 & 54.5/54.3 & 22.4/28.4 & 56.9/56.2 & 64.5/88.6 & 73.4/94.1 \\
LLaVA-OneVision-7B     & 88.5/87.8 & 51.1/55.3 & 48.6/52.6 & 25.9/33.7 & 55.3/55.4 & 52.1/79.5 & 75.1/89.1 \\
InternVL3-8B           & 88.3/89.8 & 63.4/62.2 & 51.3/50.7 & 23.6/39.0 & 54.8/50.2 & 50.8/83.1 & 63.7/87.1 \\
Llama-3.2-11B-Vision   & 90.8/89.3 & 47.7/54.0 & 45.0/45.0 & 19.0/34.5 & 50.3/50.7 & 56.8/85.7 & 60.7/91.7 \\
Qwen3.5-397B           & 96.3/96.8 & 65.5/63.7 & 68.3/69.0 & 31.8/51.8 & 70.7/69.2 & 77.3/94.8 & 73.0/97.3 \\
Gemma-4-31B            & 94.0/92.3 & 55.3/49.5 & 49.0/54.0 & 23.8/41.8 & 62.8/57.8 & 57.0/82.7 & 67.7/91.7 \\
\bottomrule
\end{tabular}%
}
\end{table}

\section{Aspect Tagger and Its Validation}
\label{sec:app-tagger}

Aspect coverage is measured by a keyword tagger rather than a model: for each
of the seven aspects we hand-wrote a list of $15$--$25$ case-insensitive
regular expressions, and an aspect counts as covered if any of its patterns
matches the critique. Table~\ref{tab:tagger-terms} lists the terms. The tagger
is applied identically to model and human critiques, so it is a consistent
instrument, but it is a shallow one: it recognizes the vocabulary in its lists
and nothing else.

\vspace{0.2cm}
Table~\ref{tab:tagger-validation} therefore places it beside an independent
measurement of the same quantity. The LLM judge of
Table~\ref{tab:aspect-engagement-judge} rates every aspect of every critique
on a $0$--$3$ scale in which $0$ means \emph{not mentioned}, so treating
scores $\geq 1$ as coverage yields a second, model-based reading of the same
$200$ posts and $4{,}790$ critiques. The two instruments agree on $79.6\%$ of
the $33{,}530$ critique--aspect decisions.

\vspace{0.2cm}
Where they disagree, they disagree in opposite directions, and inspecting the
judge's written explanations shows why. The tagger is conservative: it misses
plain-language critiques such as \hcrit{It is a bit tight on the left side of
the image. Pan to the left a bit so that the butterfly is facing into the
center}, which discusses composition without using any word on the list. The
judge is liberal: it scores the same critique as covering \emph{color} because
the critic wrote ``nice monarch''. Neither is ground truth, and the true
coverage of a critique lies between them. This matters most for human
critiques, which are shorter and less technical: the judge places human
coverage above the tagger on every aspect, most sharply for mood
($12.8\%$ against $49.1\%$) and subject ($35.8\%$ against $64.4\%$), whereas
for the models the two measures are close on most aspects.

\vspace{0.2cm}
The comparison in Section~\ref{sec:results} survives either reading. Under
both instruments every model covers composition, lighting, color, mood, and
subject more often than humans do, usually by a wide margin. The two measures
differ about the size of the gap, and about \emph{exposure}, where the judge
places LLaVA-1.6-Mistral-7B ($28.4\%$) essentially level with humans
($27.2\%$). We therefore read the coverage numbers in
Table~\ref{tab:aspect-coverage} as an upper bound on the model--human gap
rather than as a calibrated measurement of it.

\begin{table}[t]
\centering
\caption{Keyword taxonomy used by the aspect tagger. Patterns are
case-insensitive regular expressions; \texttt{.} matches any separator, so
\texttt{black.and.white} covers both hyphenated and spaced forms.}
\label{tab:tagger-terms}
\resizebox{0.9\textwidth}{!}{%
\footnotesize
\begin{tabular}{p{0.14\textwidth}p{0.60\textwidth}}
\toprule
Aspect & Patterns \\
\midrule
Composition & composition, framing, frame, rule of thirds, leading line, symmetr, balance, foreground, background, depth, perspective, crop, angle, negative space, diagonal, centered, off.center, horizon, layering, overlap, visual weight, placement, arrange \\
\addlinespace
Lighting & lighting, light, shadow, highlight, backlit, rim light, natural light, artificial light, sunlight, golden hour, flash, lamp, illuminat, silhouette, hard light, soft light, diffuse, direction of light \\
\addlinespace
Color & color, colour, hue, saturation, tone, tint, vibrant, muted, warm, cool, palette, monochrome, black.and.white, b\&w, sepia, faded, desaturat, colorful, colour grading, color grading \\
\addlinespace
Exposure & exposure, overexposed, underexposed, blown.out, clipped, histogram, stop, EV, dynamic range, highlights clipped, shadows clipped, correctly exposed, well.exposed \\
\addlinespace
Technique & sharpness, sharp, focus, depth of field, dof, bokeh, blur, noise, grain, ISO, aperture, shutter, f.stop, lens, vignette, chromatic aberration, distortion, motion blur, long exposure, handheld, tripod, post.process, edit, retouch \\
\addlinespace
Mood & mood, emotion, feeling, atmosphere, storytelling, story, narrative, impact, evocative, melanchol, joyful, serene, dramatic, tension, energy, vibe, intimacy, powerful, compelling, captivat, engage \\
\addlinespace
Subject & subject, portrait, landscape, still life, animal, wildlife, street, architecture, object, focal point, eye contact, pose, expression, person, people, model, face, figure, character \\
\bottomrule
\end{tabular}%
}
\end{table}

\section{Qualitative Examples}
\label{sec:app-examples}

\begin{figure*}[t]
\centering
\begin{examplebox}{Example 1 --- Monarch Butterfly (\texttt{7gg9b4})}
\begin{minipage}[t]{0.22\linewidth}
    \centering
    \includegraphics[width=\linewidth]{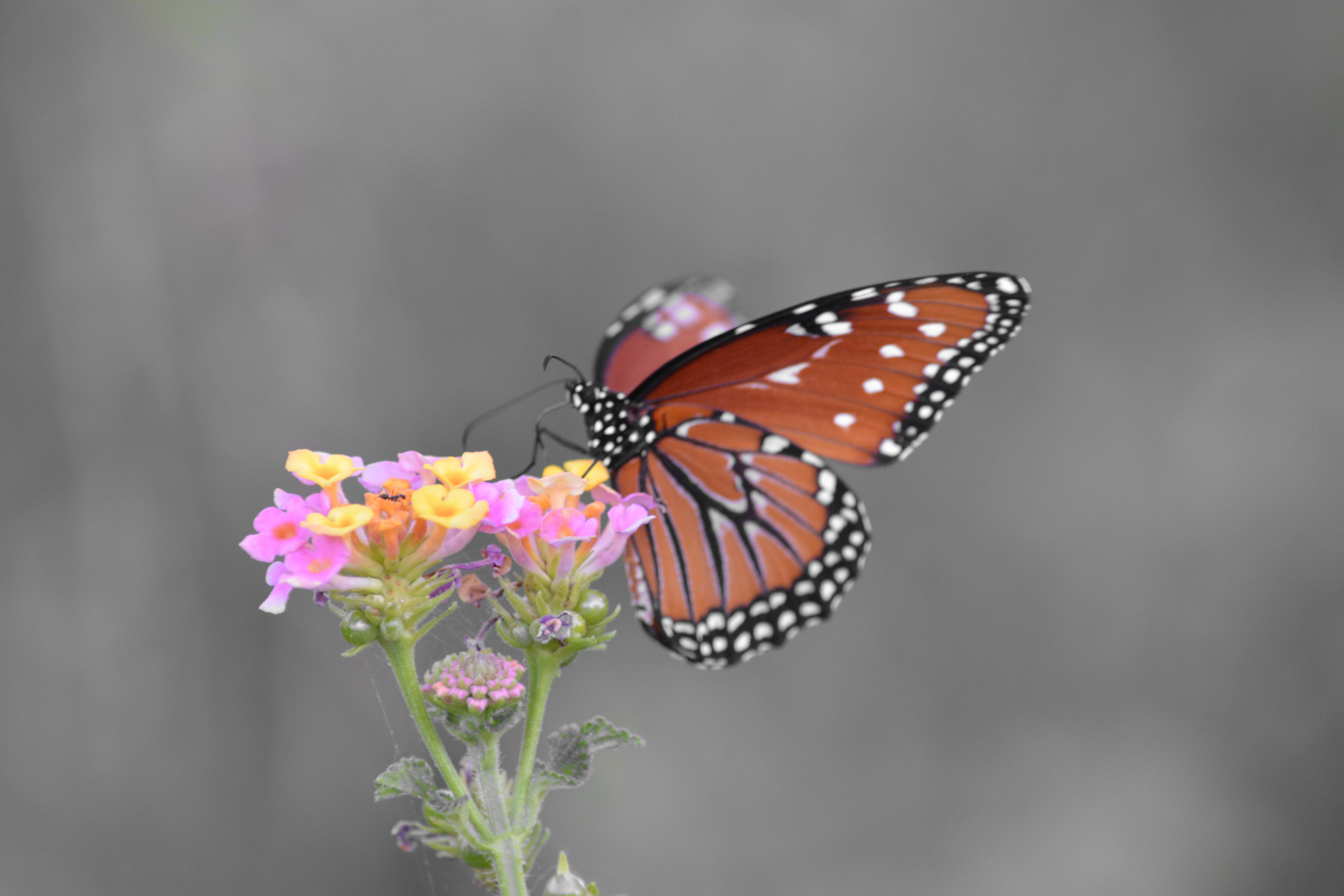}
    \vspace{2pt}
    \small\textbf{Post:} \textit{``Besides the fact that the butterfly isnt in focus, how is the overall composition? How can this image be improved?''}
\end{minipage}
\hfill
\begin{minipage}[t]{0.74\linewidth}
    \exlabel{Human critiques}{exhumanbg}
    \hcrit{Nice monarch! It is a bit tight on the left side of the image. Pan to the
    left a bit so that the butterfly is facing into the center of the image, instead
    of the border. Hope this helps!} \textsc{[39w]}

    \smallskip
    \hcrit{I would crop some of the dead space. One question - is the background
    naturally grey or did you desaturated it to that extent? If it was like that, I
    would add some color, slight green tint. The grey background doesn't look good
    at all.} \textsc{[45w]}

    \exlabel{MLLM critiques (\textsc{Capped-Plain})}{exvlmbg}
    \textbf{LLaVA-1.6-Mistral-7B} \textsc{[157w]}: ``As an experienced photographer, I
    appreciate the effort put into capturing this image of a butterfly perched on a
    flower. However, there are several areas for improvement [\ldots] Firstly, the
    butterfly itself appears to be out of focus, which detracts from the sharpness of
    the subject. To address this issue, the photographer could have used a faster
    shutter speed or increased the depth of field [\ldots]''

    \smallskip
    \textbf{Llama-3.2-11B} \textsc{[125w]}: ``The composition of this image is strong,
    with the butterfly positioned centrally and the flower placed in the foreground.
    [\ldots] However, the lack of focus on the butterfly is a significant issue, as it
    detracts from the overall impact of the image. To improve the image, I would
    recommend re-shooting with a faster shutter speed to freeze the butterfly in
    motion [\ldots]''

    \smallskip
    \textbf{Qwen3.5-397B} \textsc{[140w]}: ``You're right to flag the focus issue --- it's
    critical in macro work. [\ldots] Compositionally, you've got a strong diagonal
    flow from flower to wing tip, which guides the eye well. However, the gray
    background feels flat and disconnected [\ldots] Also, the flower cluster is
    slightly cropped at the bottom --- give it more breathing room or reframe tighter
    on the butterfly alone.''
\end{minipage}
\end{examplebox}
\caption{Example 1 (post \texttt{7gg9b4}, monarch butterfly). The poster sets the focus
problem aside and asks about composition. Both 7--11B models lead with the excluded
focus problem anyway, and Llama-3.2-11B praises the central framing that both humans
advise changing. The open-weight frontier Qwen3.5-397B answers the question that was
asked and converges with the humans on the flat gray background and the tight crop.}
\label{fig:example1}
\end{figure*}

\begin{figure*}[t]
\centering
\begin{examplebox}{Example 2 --- Night Street Photography (\texttt{646iay})}
\begin{minipage}[t]{0.22\linewidth}
    \centering
    \includegraphics[width=\linewidth]{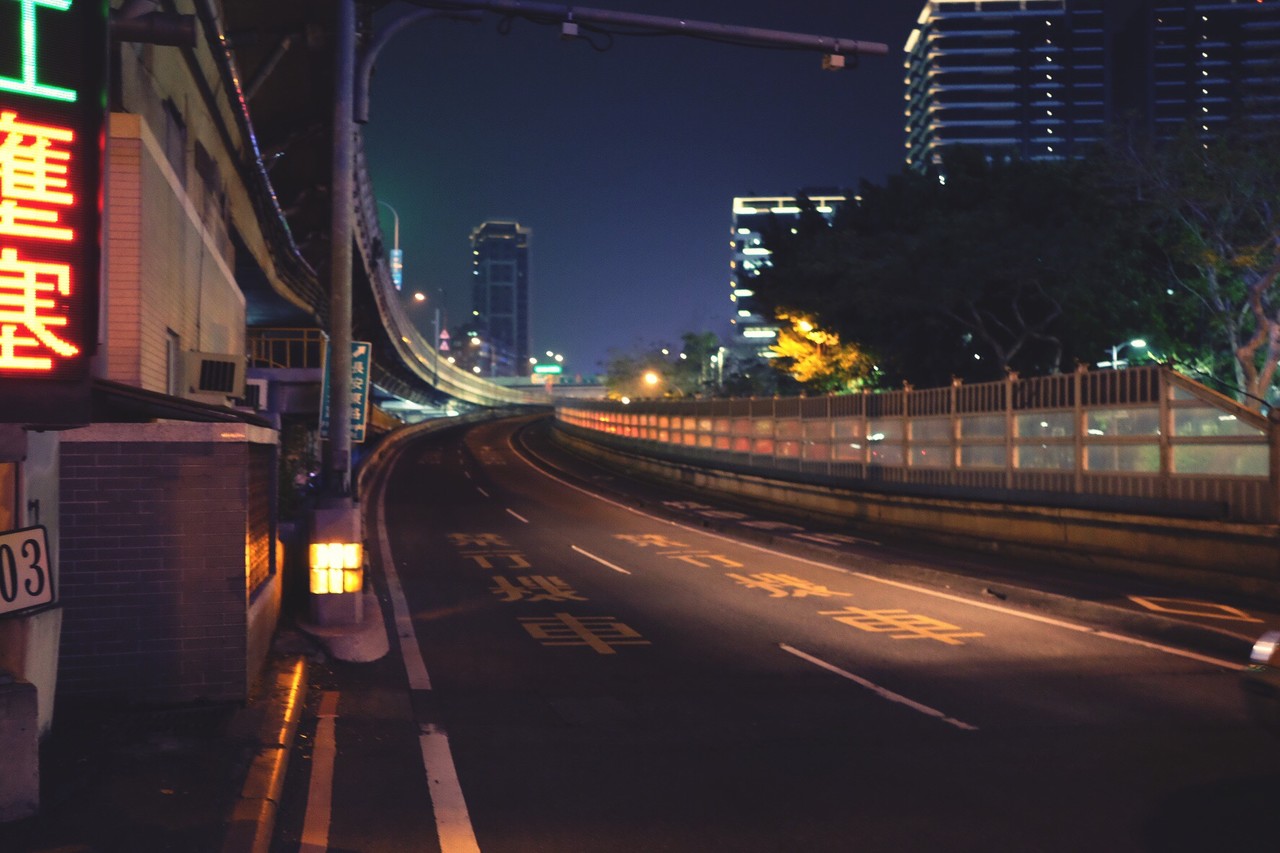}
    \vspace{2pt}
    \small\textbf{Post:} \textit{``Street in focus, everything else blurry. Should I have focused another part?''}
\end{minipage}
\hfill
\begin{minipage}[t]{0.74\linewidth}
    \exlabel{Human critiques}{exhumanbg}
    \hcrit{Your aperture is what's causing most of it. I would use a tripod if I were
    doing night photography, because it lets me use a slower shutter speed without
    motion blur. [\ldots] Another thing you can try to do if you want everything in
    focus is use hyperfocal distance.} \textsc{[77w]}

    \smallskip
    \hcrit{An image like this, with no strong subject, relies heavily on composition.
    The sign at the left should not be cropped, the light pole horizontal needs to be
    square to the frame to show deliberate control. The trees at middle right ruin
    this photo more than anything else, they block the buildings, disrupting what
    could have been a strong geometric study.} \textsc{[84w]}

    \exlabel{MLLM critiques (\textsc{Capped-Plain})}{exvlmbg}
    \textbf{InternVL3-8B} \textsc{[120w]}: ``This image captures a striking urban night
    scene, with the main road sharply in focus and an intentional blur applied to the
    background. The sharpness draws attention effectively to the street details and
    signage, creating depth through selective focus. [\ldots] Overall, the composition
    is strong, emphasizing the foreground while maintaining context through subtle
    lighting from surrounding buildings.''

    \smallskip
    \textbf{LLaVA-OneVision-7B} \textsc{[179w]}: ``The photograph presents an
    interesting play of light and focus that creates a sense of depth and atmosphere.
    [\ldots] This technique can be quite effective for emphasizing the subject or for
    artistic purposes. [\ldots] If the goal was to create a sense of isolation or
    solitude, keeping the background blurred works well.''

    \smallskip
    \textbf{Kimi-K2.6} \textsc{[99w]}: ``The focus choice undermines your strongest
    element. That neon sign is begging to be sharp---it's where the eye goes first, yet
    it's soft. An empty road, even in focus, doesn't compete with that glow and color.
    I'd have either focused on the sign or stopped down to f/8--f/11 to get both sign
    and street sharp. [\ldots] The mood and color are great, but right now the blur
    feels accidental rather than intentional.''
\end{minipage}
\end{examplebox}
\caption{Example 2 (post \texttt{646iay}, night street). The poster is unsure whether
the pervasive blur was a mistake. Humans diagnose its causes (aperture, tripod,
hyperfocal distance) and name concrete compositional flaws. The 7--8B models describe
the same blur as effective selective focus that creates depth; the frontier Kimi-K2.6
diagnoses it as the humans do, naming the missed subject and the aperture fix.}
\label{fig:example2}
\end{figure*}

\begin{figure*}[t]
\centering
\begin{examplebox}{Example 3 --- Wood-Burned Art Product Photo (\texttt{ayld0g})}
\begin{minipage}[t]{0.22\linewidth}
    \centering
    \includegraphics[width=\linewidth]{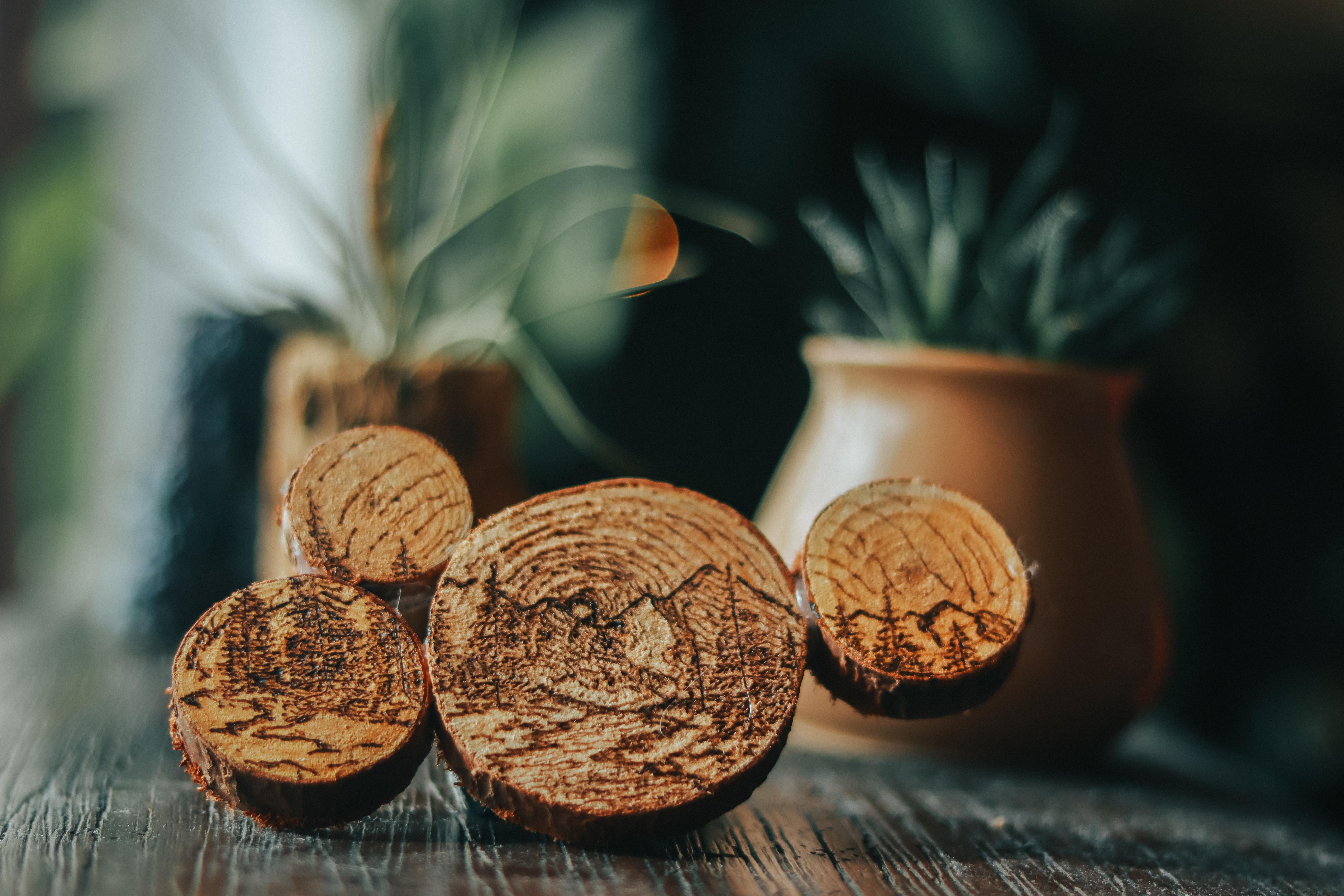}
    \vspace{2pt}
    \small\textbf{Post:} \textit{``Photo I took for a buddy to sell her wood burning art''}
\end{minipage}
\hfill
\begin{minipage}[t]{0.74\linewidth}
    \exlabel{Human critiques}{exhumanbg}
    \hcrit{Awesome start to your product photography career! The lighting looks quite
    natural and the dark green in the background complements the product. As for
    feedback, I do think the background is a little distracting overall - the right
    side is great, but the left is too busy. Also don't be afraid to stop your lens
    down to get more sharpness and depth of field.} \textsc{[80w]}

    \smallskip
    \hcrit{I think the aperture is slightly too narrow. The tops of the wood burnings
    are going out of focus. I'd do 2/3 of a stop or a stop wider. I like the green
    environment you've put it in. It creates a nice nature vibe.} \textsc{[44w]}

    \exlabel{MLLM critiques (\textsc{Capped-Plain})}{exvlmbg}
    \textbf{Qwen2-VL-7B} \textsc{[126w]}: ``The image effectively showcases the
    intricate details of the woodburned pieces, highlighting the craftsmanship
    involved. The shallow depth of field draws attention to the foreground while softly
    blurring the background, creating a pleasing bokeh effect that adds visual interest
    without distracting from the main subject. [\ldots] Overall, it's a strong
    representation of the art, but refining the technical aspects would enhance its
    quality further.''

    \smallskip
    \textbf{Gemma-4-31B} \textsc{[117w]}: ``Nice shot! You've captured a great mood
    here, and the shallow depth of field does a good job of separating the art from the
    background. However, since this is for a product sale, the focus is a bit too soft.
    The center piece is slightly out of focus, and the edges blur quickly. [\ldots] I'd
    suggest stopping down your aperture (increase your f-stop) to get more of the
    foreground in focus.''

    \smallskip
    \textbf{GPT-5.5} \textsc{[96w]}: ``Nice atmosphere---the warm tones, shallow depth
    of field, and natural setting fit the handmade/woodcraft feel well. For a sales
    photo, though, the product itself needs to be clearer. The focus feels a bit soft
    on the burned designs, and the strong blur/dark background competes with
    readability. I'd shoot a version at a slightly narrower aperture (f/5.6-ish) so all
    pieces are sharp [\ldots]''
\end{minipage}
\end{examplebox}
\caption{Example 3 (post \texttt{ayld0g}, product photo of wood-burned art). Every human
wants more depth of field so the product reads sharply. Qwen2-VL-7B praises the shallow
depth of field as a pleasing bokeh effect; the frontier models side with the humans,
naming the same fix and concrete settings.}
\label{fig:example3}
\end{figure*}

\begin{figure*}[t]
\centering
\begin{examplebox}{Example 4 --- Dark Trailer Scene (\texttt{8kfica})}
\begin{minipage}[t]{0.22\linewidth}
    \centering
    \includegraphics[width=\linewidth]{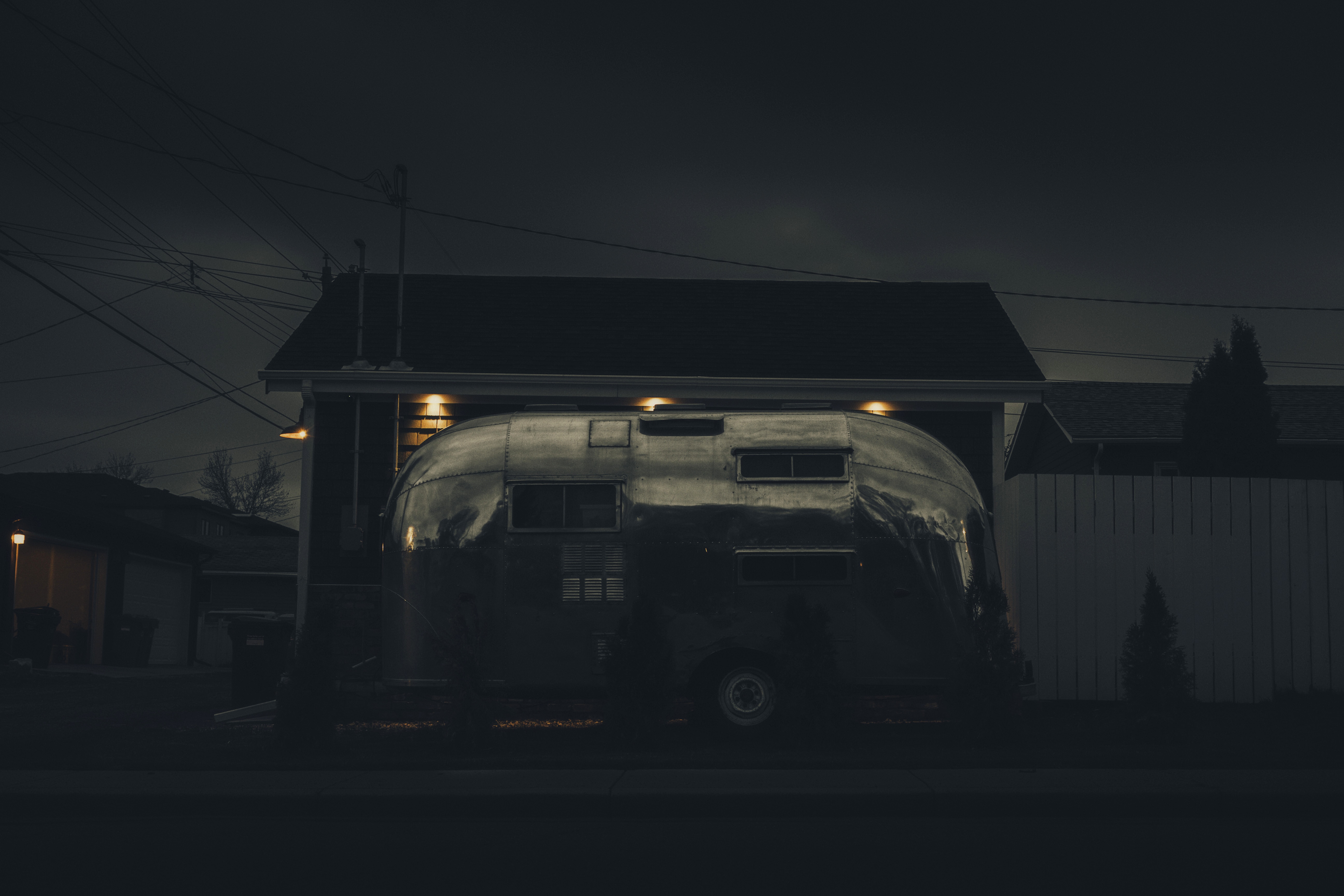}
    \vspace{2pt}
    \small\textbf{Post:} \textit{``I like to shoot dark and moody scenes. What do you think about this, is it too dark?''}
\end{minipage}
\hfill
\begin{minipage}[t]{0.74\linewidth}
    \exlabel{Human critiques}{exhumanbg}
    \hcrit{yes, a little too dark. also would not look good on any screen where the
    brightness is not set to max} \textsc{[21w]}

    \smallskip
    \hcrit{It's a little too dark for me, primarily because it's really hard to see
    any detail in the bottom fourth of the image. I would raise the blacks juuuuust a
    little bit. [\ldots]} \textsc{[82w]}

    \smallskip
    \hcrit{I don't think it is too dark, but I think it's important how it's
    presented. The way we're currently viewing it, on Reddit, matted by white bars on
    either side, may create a false impression of it being too dark.} \textsc{[40w]}

    \exlabel{MLLM critiques (\textsc{Capped-Plain})}{exvlmbg}
    \textbf{Qwen2-VL-7B} \textsc{[83w]}: ``The image effectively captures a dark and
    moody atmosphere, which alignss well with the photographer's style. The lighting
    focuses on the metallic trailer, creating a dramatic contrast against the darker
    surroundings. However, there might be some areas that could benefit from slight
    adjustments for better balance. [\ldots] Overall, the composition is strong, but
    refining these details could enhance the visual impact further.''

    \smallskip
    \textbf{Gemma-4-31B} \textsc{[129w]}: ``The composition is strong, and the
    Airstream is a fantastic subject for this aesthetic. To answer your question: yes,
    it is a bit too dark. While `moody' is the goal, you've lost critical detail in
    the shadows [\ldots] I'd suggest lifting the shadows slightly in post-processing
    to define the trailer's silhouette against the house.''

    \smallskip
    \textbf{GPT-5.5} \textsc{[115w]}: ``Not too dark for the mood you're aiming
    for---the low-key exposure gives the scene a nice cinematic, slightly eerie feel.
    [\ldots] That said, the lower third is very close to empty black, so the image
    loses some structure there. I'd consider lifting the shadows just slightly on the
    trailer/wheel area [\ldots]''
\end{minipage}
\end{examplebox}
\caption{Example 4 (post \texttt{8kfica}, dark trailer scene). A question with no single
right answer: the humans split two to one, and the frontier models split the same way,
Gemma-4-31B answering yes and GPT-5.5 no, while both converge with the second human on
the identical remedy of lifting the shadows slightly. The small model does not answer
the question, praising the mood instead.}
\label{fig:example4}
\end{figure*}

\textbf{Qualitative failure modes.}
Figures~\ref{fig:example1}--\ref{fig:example4} show four posts from the
benchmark, each with the human references and \textsc{Capped-Plain} critiques
from three models; all nine models of the study appear across the four
figures. The examples make concrete the failure modes whose prevalence is
quantified in Section~\ref{sec:results}. The first is
\emph{question-insensitivity}: a poster explicitly sets the focus problem
aside and asks about composition, yet the 7--11B models lead with the focus
problem anyway and run through a standard set of aspects
(Fig.~\ref{fig:example1}), and a poster asking whether the scene is too dark
receives praise for its mood instead of an answer
(Fig.~\ref{fig:example4}). The second is the \emph{reframing of flaws as
intentional choices}: the poster is unsure whether the pervasive blur was a
mistake, the humans diagnose its cause, and the small models describe the
same blur as effective selective focus that creates depth
(Fig.~\ref{fig:example2}). The third is \emph{praising the flaw}: every
human wants more depth of field so the product reads sharply, and the small
model presents the shallow depth of field as a pleasing bokeh effect
(Fig.~\ref{fig:example3}).

\vspace{0.2cm}
\textbf{Agreement at the frontier.}
The frontier models behave differently, and the examples show both halves.
In each case Qwen3.5-397B, Gemma-4-31B, Kimi-K2.6 or GPT-5.5 answers the
question that was asked, diagnoses the defect, and converges with the human
advice down to the concrete remedy: crop the dead space, stop down the
aperture, give the frame a subject, lift the shadows. This is consistent
with the quantitative picture, in which image dependence rises with
capability in the grounding test and the LLM judges prefer frontier
critiques over the human references in a majority of pairs.
Figure~\ref{fig:example4} adds the complementary observation: where the
humans themselves disagree, splitting two to one on whether the scene is too
dark, the two frontier models split the same way while agreeing on the
remedy. Competent critics disagreeing about a defensible question is the
spread that motivates the multiple-reference design of
Section~\ref{sec:setup}.

\vspace{0.2cm}
\textbf{What the examples share.}
What does not change with scale is the register. Every model critique in the
four figures runs $83$ to $179$ words against shown human references of $21$
to $84$, works through most of the aspect list whatever was asked, and wraps
its criticism in a reliably encouraging frame. This shared register is
exactly the property that inflates the embedding-cosine similarity in
Section~\ref{sec:results} while the stricter, meaning-sensitive metrics stay
low: fluent, comprehensive, affirmative critique that overlaps in vocabulary
with the human references without matching their selectivity.

\clearpage

\clearpage
\begin{figure*}[t]
\centering
\includegraphics[width=.9\textwidth]{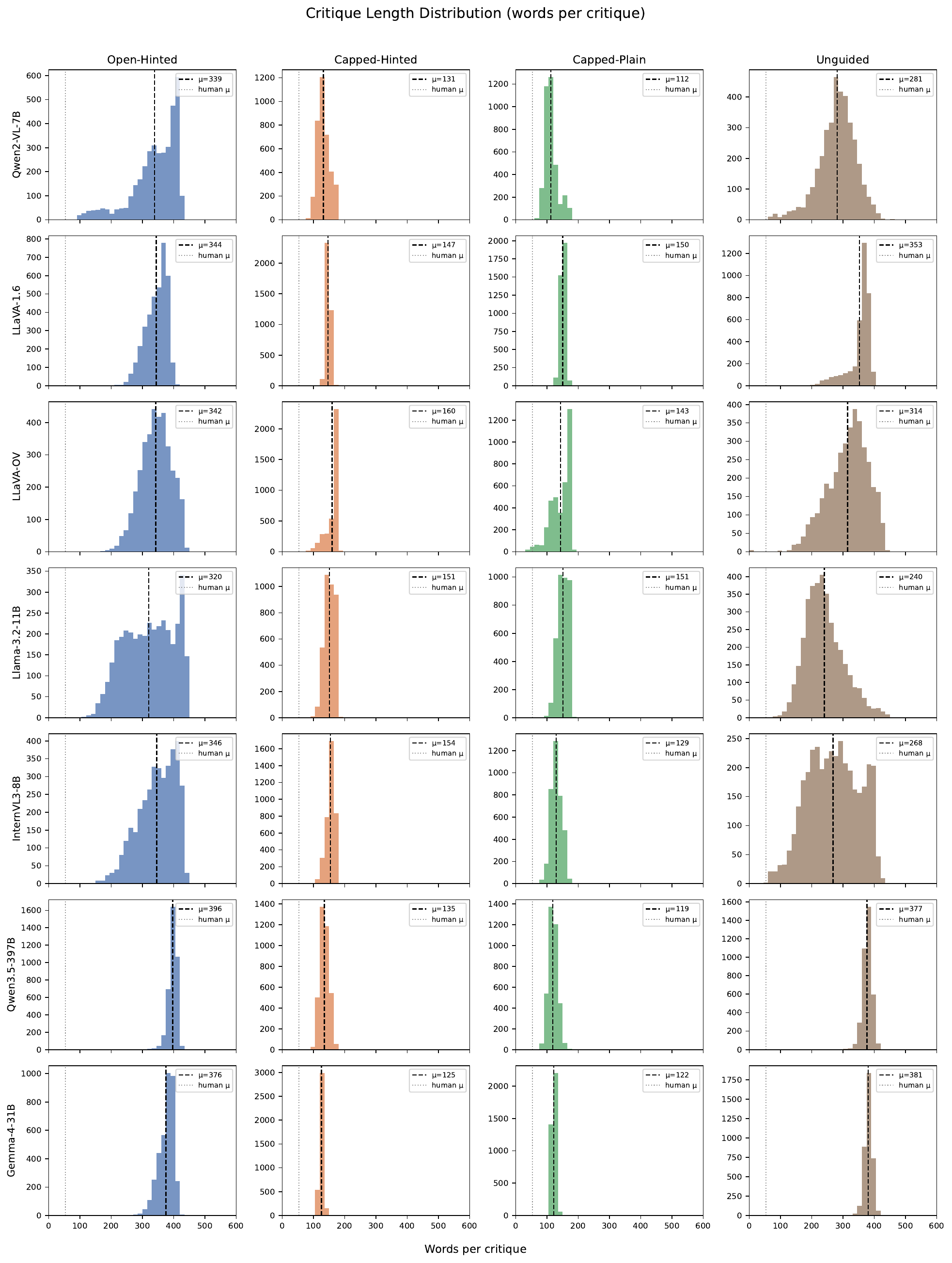}
\caption{Critique length under the four conditions that request a single
critique: seven models (rows) by four prompt conditions (columns). The dashed
line marks each model's mean ($\mu$, shown in each panel); the dotted line
marks the human mean of $54.0$ words.}
\label{fig:length-hist}
\end{figure*}

\begin{figure*}[t]
\centering
\includegraphics[width=0.9\textwidth]{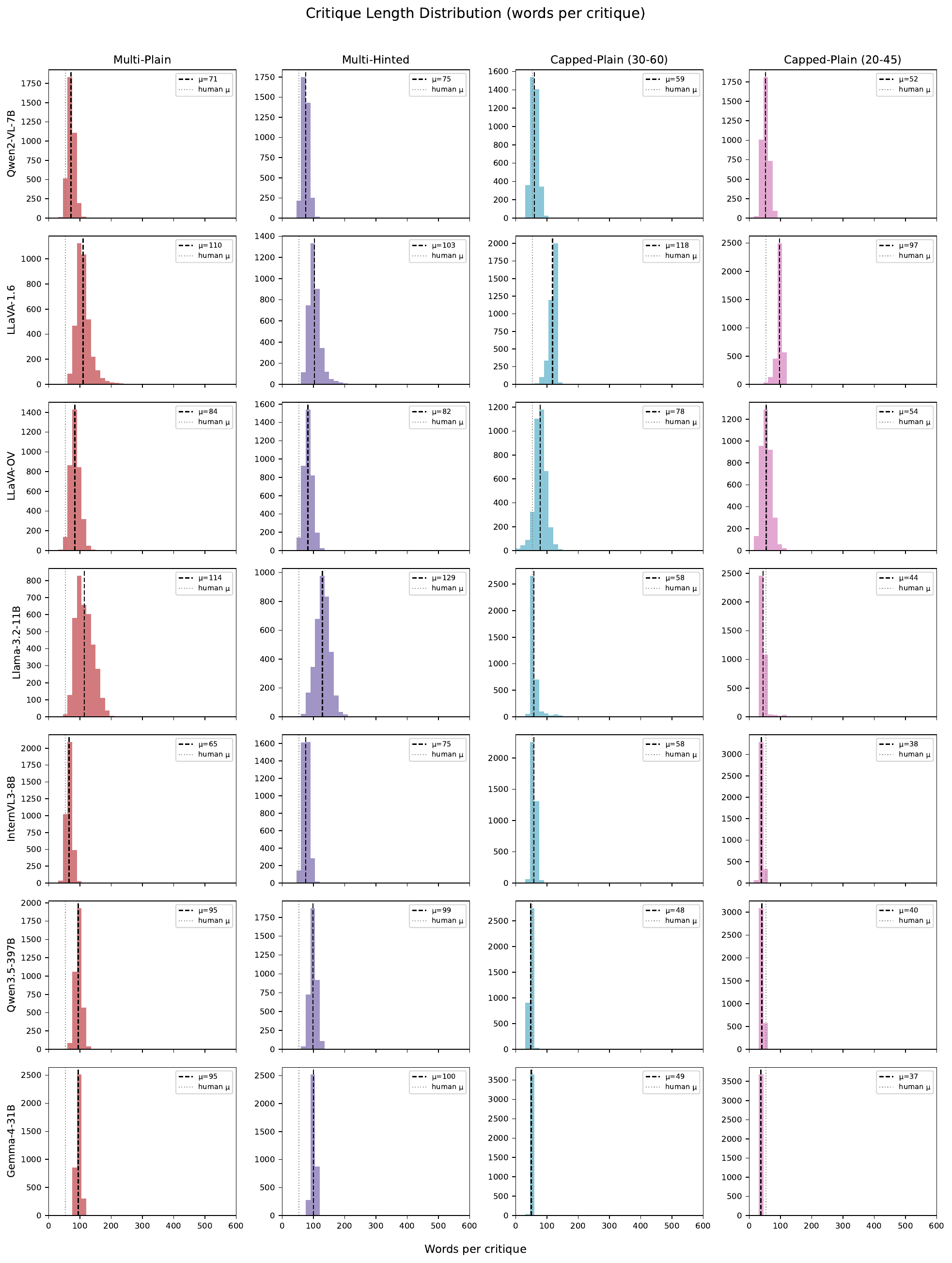}
\caption{Critique length under the remaining four conditions: the two that
request three critiques in a single pass and the two length-matched variants
of \textsc{Capped-Plain}. Lines are as in Fig.~\ref{fig:length-hist}.}
\label{fig:length-hist-capped}
\end{figure*}

\section{Critique Length Distributions}
\label{sec:app-length}

Figures~\ref{fig:length-hist} and~\ref{fig:length-hist-capped} give the full
per-model, per-condition length distributions behind the length analysis of
Section~\ref{sec:results}: seven models by four conditions in each figure,
with the human mean of $54.0$ words marked in every panel. Three regimes are
visible. Without a stated word budget the models write three to seven times
human length (\textsc{Open-Hinted} $320$--$397$ words, \textsc{Unguided}
$240$--$381$). With a $50$--$150$ budget they comply but settle near its
upper bound (\textsc{Capped-Hinted} $125$--$160$, \textsc{Capped-Plain}
$112$--$151$) rather than near the human mean, and splitting the output into
three critiques shortens each one without any instruction to be brief
(\textsc{Multi-Plain} $65$--$114$, \textsc{Multi-Hinted} $75$--$129$). Only
the two length-matched conditions bring the models into the human range:
\textsc{Capped-Plain} $(30$--$60)$ yields $47.6$--$117.7$ words and
$(20$--$45)$ yields $37.1$--$96.6$. Models can therefore write at human
length when the range is stated tightly, so the overshoot everywhere else
reflects calibration rather than an inability to be concise.

\vspace{0.2cm}
\textbf{Word-limit compliance.} The word range in each prompt is a
natural-language instruction, not hard truncation, and compliance is partial
(Table~\ref{tab:compliance}). Counting whitespace-delimited tokens on the raw
generation, $80.1\%$ of \textsc{Capped-Plain} critiques fall inside the
requested $50$--$150$ words and $63.3\%$ under \textsc{Capped-Hinted}, with
every violation an overshoot; none fall below the lower bound. Compliance
tracks capability rather than the prompt: Gemma-4-31B, Kimi-K2.6 and GPT-5.5
comply on every critique, whereas the two LLaVA models and
Llama-3.2-11B-Vision comply on fewer than half.

\vspace{0.2cm}
The decoding budget of Section~\ref{sec:setup} is a second, harder
constraint. For the capped conditions it is $200$ tokens, which for these
models corresponds to roughly $175$--$190$ words, and the verbose models run
into it: their critiques stop mid-sentence, which we detect as the absence of
terminal punctuation, on $21.2\%$ (Llama-3.2-11B-Vision), $37.5\%$
(LLaVA-OneVision-7B) and $92.1\%$ (LLaVA-1.6-Mistral-7B) of critiques, and
their length distributions terminate abruptly at $173$--$188$ words. The
models that respect the requested range never approach the budget, topping
out at $141$--$152$ words with no truncation. Reported lengths for the three
verbose models are therefore censored, and the overshoot in
Fig.~\ref{fig:length-hist} is a lower bound on what they would otherwise
write.

\begin{table}[t]
\centering
\caption{Word-limit compliance under the two $50$--$150$ word conditions
($N{=}1{,}227$, $3{,}681$ critiques per model). \emph{In range} is the share
of critiques inside the requested range; \emph{unterm.} the share ending
without terminal punctuation, our truncation signature; \emph{max} the
longest critique observed, where the $200$-token budget binds. Kimi-K2.6 and
GPT-5.5 were run in the primary condition only.}
\label{tab:compliance}
\resizebox{0.8\textwidth}{!}{%
\small
\begin{tabular}{lccccc}
\toprule
& \multicolumn{3}{c}{Capped-Plain} & \multicolumn{2}{c}{Capped-Hinted} \\
\cmidrule(lr){2-4} \cmidrule(lr){5-6}
Model & In range (\%) & Unterm. (\%) & Max & In range (\%) & Unterm. (\%) \\
\midrule
Qwen2-VL-7B            & 91.3 & 7.3 & 188 & 81.7 & 10.4 \\
LLaVA-1.6-Mistral-7B   & 49.6 & 92.1 & 173 & 71.6 & 92.1 \\
LLaVA-OneVision-7B     & 46.6 & 37.5 & 186 & 22.1 & 66.2 \\
InternVL3-8B           & 86.8 & 8.8 & 180 & 33.5 & 36.7 \\
Llama-3.2-11B-Vision   & 47.9 & 21.2 & 185 & 48.8 & 21.7 \\
Qwen3.5-397B           & 98.5 & 12.9 & 168 & 85.2 & 10.0 \\
Gemma-4-31B            & 100.0 & 0.1 & 142 & 100.0 & 0.0 \\
Kimi-K2.6              & 100.0 & 0.0 & 152 & -- & -- \\
GPT-5.5                & 100.0 & 0.0 & 141 & -- & -- \\
\bottomrule
\end{tabular}%
}
\end{table}

\clearpage

\begin{table}[t]
\centering
\footnotesize
\setlength{\tabcolsep}{4.5pt}
\caption{Critique--human similarity for all seven open-weight models under
the six prompt conditions of the main study ($N{=}1{,}227$, 3$\times$3
max-mapping). \emph{Words} is the mean critique length. \emph{Human} is the
leave-one-out agreement baseline among the three human critiques per post.}
\label{tab:full-similarity}
\begin{tabular}{lrrrrr}
\toprule
Condition & Words & ROUGE-L & SBERT-cos & BERTScore & BLEURT \\
\midrule
\multicolumn{6}{l}{\textbf{Qwen2-VL-7B}} \\
Unguided               & 280.8 & 0.110 & 0.624 & $-0.0303$ & $-0.750$ \\
Open-Hinted            & 339.1 & 0.101 & 0.611 & $-0.0567$ & $-0.778$ \\
Capped-Hinted          & 131.5 & 0.137 & 0.594 & $\phantom{-}0.0735$ & $-0.608$ \\
Capped-Plain           & 112.4 & 0.142 & 0.621 & $\phantom{-}0.0986$ & $-0.601$ \\
Multi-Hinted           &  75.2 & 0.151 & 0.531 & $\phantom{-}0.1125$ & $-0.551$ \\
Multi-Plain            &  71.1 & 0.154 & 0.587 & $\phantom{-}0.1395$ & $-0.514$ \\
\addlinespace
\multicolumn{6}{l}{\textbf{LLaVA-1.6-Mistral-7B}} \\
Unguided               & 352.8 & 0.099 & 0.612 & $-0.0481$ & $-0.779$ \\
Open-Hinted            & 344.1 & 0.101 & 0.588 & $-0.0473$ & $-0.801$ \\
Capped-Hinted          & 146.7 & 0.133 & 0.557 & $\phantom{-}0.0384$ & $-0.755$ \\
Capped-Plain           & 150.1 & 0.134 & 0.594 & $\phantom{-}0.0548$ & $-0.698$ \\
Multi-Hinted           & 103.2 & 0.144 & 0.553 & $\phantom{-}0.0982$ & $-0.558$ \\
Multi-Plain            & 110.1 & 0.145 & 0.612 & $\phantom{-}0.1079$ & $-0.535$ \\
\addlinespace
\multicolumn{6}{l}{\textbf{LLaVA-OneVision-7B}} \\
Unguided               & 314.4 & 0.104 & 0.615 & $-0.0337$ & $-0.722$ \\
Open-Hinted            & 342.2 & 0.099 & 0.603 & $-0.0435$ & $-0.704$ \\
Capped-Hinted          & 159.5 & 0.127 & 0.577 & $\phantom{-}0.0404$ & $-0.638$ \\
Capped-Plain           & 143.0 & 0.132 & 0.611 & $\phantom{-}0.0655$ & $-0.631$ \\
Multi-Hinted           &  82.4 & 0.143 & 0.537 & $\phantom{-}0.0998$ & $-0.585$ \\
Multi-Plain            &  83.9 & 0.147 & 0.594 & $\phantom{-}0.1189$ & $-0.561$ \\
\addlinespace
\multicolumn{6}{l}{\textbf{InternVL3-8B}} \\
Unguided               & 267.9 & 0.104 & 0.616 & $-0.0542$ & $-0.797$ \\
Open-Hinted            & 345.9 & 0.093 & 0.611 & $-0.0636$ & $-0.767$ \\
Capped-Hinted          & 153.9 & 0.122 & 0.592 & $\phantom{-}0.0433$ & $-0.609$ \\
Capped-Plain           & 129.2 & 0.128 & 0.615 & $\phantom{-}0.0736$ & $-0.602$ \\
Multi-Hinted           &  75.4 & 0.133 & 0.535 & $\phantom{-}0.0893$ & $-0.640$ \\
Multi-Plain            &  64.8 & 0.138 & 0.603 & $\phantom{-}0.1276$ & $-0.591$ \\
\addlinespace
\multicolumn{6}{l}{\textbf{Llama-3.2-11B-Vision}} \\
Unguided               & 240.0 & 0.121 & 0.599 & $-0.0026$ & $-0.697$ \\
Open-Hinted            & 320.1 & 0.107 & 0.608 & $-0.0351$ & $-0.686$ \\
Capped-Hinted          & 150.7 & 0.136 & 0.594 & $\phantom{-}0.0690$ & $-0.616$ \\
Capped-Plain           & 150.7 & 0.141 & 0.626 & $\phantom{-}0.0834$ & $-0.619$ \\
Multi-Hinted           & 128.9 & 0.135 & 0.584 & $\phantom{-}0.0661$ & $-0.644$ \\
Multi-Plain            & 114.2 & 0.151 & 0.619 & $\phantom{-}0.1246$ & $-0.577$ \\
\addlinespace
\multicolumn{6}{l}{\textbf{Qwen3.5-397B-A17B}} \\
Unguided               & 376.9 & 0.096 & 0.635 & $-0.0824$ & $-0.924$ \\
Open-Hinted            & 396.5 & 0.095 & 0.644 & $-0.0510$ & $-0.927$ \\
Capped-Hinted          & 134.7 & 0.131 & 0.634 & $\phantom{-}0.0684$ & $-0.793$ \\
Capped-Plain           & 118.8 & 0.135 & 0.641 & $\phantom{-}0.0816$ & $-0.852$ \\
Multi-Hinted           &  98.6 & 0.136 & 0.582 & $\phantom{-}0.0744$ & $-0.810$ \\
Multi-Plain            &  94.5 & 0.138 & 0.633 & $\phantom{-}0.0960$ & $-0.856$ \\
\addlinespace
\multicolumn{6}{l}{\textbf{Gemma-4-31B}} \\
Unguided               & 380.5 & 0.101 & 0.646 & $-0.0731$ & $-0.937$ \\
Open-Hinted            & 375.5 & 0.104 & 0.645 & $-0.0275$ & $-0.945$ \\
Capped-Hinted          & 125.2 & 0.146 & 0.648 & $\phantom{-}0.1040$ & $-0.837$ \\
Capped-Plain           & 121.5 & 0.147 & 0.648 & $\phantom{-}0.1031$ & $-0.876$ \\
Multi-Hinted           &  99.6 & 0.149 & 0.586 & $\phantom{-}0.1020$ & $-0.739$ \\
Multi-Plain            &  94.6 & 0.159 & 0.638 & $\phantom{-}0.1376$ & $-0.745$ \\
\midrule
\textbf{Human} &  54.0 & 0.147 & 0.514 & $\phantom{-}0.1139$ & $-1.077$ \\
\bottomrule
\end{tabular}
\end{table}

\clearpage

\begin{table}[t]
\centering
\footnotesize
\setlength{\tabcolsep}{4.5pt}
\caption{Effect of the output-length budget: \textsc{Capped-Plain}
($50$--$150$ words, repeated from Table~\ref{tab:full-similarity}) against
the identical prompt with only the requested word range changed
($N{=}1{,}227$).}
\label{tab:length-matched}
\begin{tabular}{lrrrrr}
\toprule
Condition & Words & ROUGE-L & SBERT-cos & BERTScore & BLEURT \\
\midrule
\multicolumn{6}{l}{\textbf{Qwen2-VL-7B}} \\
Capped-Plain (50--150) & 112.4 & 0.142 & 0.621 & $\phantom{-}0.0986$ & $-0.601$ \\
Capped-Plain (30--60)  &  59.3 & 0.155 & 0.624 & $\phantom{-}0.1571$ & $-0.561$ \\
Capped-Plain (20--45)  &  51.6 & 0.157 & 0.623 & $\phantom{-}0.1664$ & $-0.565$ \\
\addlinespace
\multicolumn{6}{l}{\textbf{LLaVA-1.6-Mistral-7B}} \\
Capped-Plain (50--150) & 150.1 & 0.134 & 0.594 & $\phantom{-}0.0548$ & $-0.698$ \\
Capped-Plain (30--60)  & 117.7 & 0.142 & 0.604 & $\phantom{-}0.0918$ & $-0.624$ \\
Capped-Plain (20--45)  &  96.6 & 0.146 & 0.601 & $\phantom{-}0.1060$ & $-0.599$ \\
\addlinespace
\multicolumn{6}{l}{\textbf{LLaVA-OneVision-7B}} \\
Capped-Plain (50--150) & 143.0 & 0.132 & 0.611 & $\phantom{-}0.0655$ & $-0.631$ \\
Capped-Plain (30--60)  &  77.9 & 0.149 & 0.608 & $\phantom{-}0.1231$ & $-0.576$ \\
Capped-Plain (20--45)  &  54.0 & 0.155 & 0.614 & $\phantom{-}0.1556$ & $-0.572$ \\
\addlinespace
\multicolumn{6}{l}{\textbf{InternVL3-8B}} \\
Capped-Plain (50--150) & 129.2 & 0.128 & 0.615 & $\phantom{-}0.0736$ & $-0.602$ \\
Capped-Plain (30--60)  &  57.6 & 0.145 & 0.615 & $\phantom{-}0.1386$ & $-0.569$ \\
Capped-Plain (20--45)  &  38.3 & 0.146 & 0.611 & $\phantom{-}0.1628$ & $-0.593$ \\
\addlinespace
\multicolumn{6}{l}{\textbf{Llama-3.2-11B-Vision}} \\
Capped-Plain (50--150) & 150.7 & 0.141 & 0.626 & $\phantom{-}0.0834$ & $-0.619$ \\
Capped-Plain (30--60)  &  58.3 & 0.158 & 0.623 & $\phantom{-}0.1710$ & $-0.529$ \\
Capped-Plain (20--45)  &  43.6 & 0.159 & 0.615 & $\phantom{-}0.1872$ & $-0.534$ \\
\addlinespace
\multicolumn{6}{l}{\textbf{Qwen3.5-397B-A17B}} \\
Capped-Plain (50--150) & 118.8 & 0.135 & 0.641 & $\phantom{-}0.0816$ & $-0.852$ \\
Capped-Plain (30--60)  &  47.6 & 0.146 & 0.624 & $\phantom{-}0.1346$ & $-0.894$ \\
Capped-Plain (20--45)  &  40.3 & 0.146 & 0.613 & $\phantom{-}0.1443$ & $-0.883$ \\
\addlinespace
\multicolumn{6}{l}{\textbf{Gemma-4-31B}} \\
Capped-Plain (50--150) & 121.5 & 0.147 & 0.648 & $\phantom{-}0.1031$ & $-0.876$ \\
Capped-Plain (30--60)  &  49.4 & 0.163 & 0.631 & $\phantom{-}0.1638$ & $-0.822$ \\
Capped-Plain (20--45)  &  37.1 & 0.163 & 0.618 & $\phantom{-}0.1751$ & $-0.833$ \\
\midrule
\textbf{Human} &  54.0 & 0.147 & 0.514 & $\phantom{-}0.1139$ & $-1.077$ \\
\bottomrule
\end{tabular}
\end{table}

\section{Full Reference-Similarity Results}
\label{sec:app-full-similarity}

Table~\ref{tab:full-similarity} extends Tables~\ref{tab:main}
and~\ref{tab:conditions} to all seven open-weight models under the six
prompt conditions of the main study; Kimi-K2.6 and GPT-5.5 were run in the
primary condition only and appear in Table~\ref{tab:main}. The patterns of
Section~\ref{sec:results} hold for every model: the uncapped conditions
score negative on BERTScore, ROUGE-L rises as critiques shorten, and
SBERT-cos stays between $0.53$ and $0.65$ throughout.

\vspace{0.2cm}
Table~\ref{tab:length-matched} isolates the output-length budget.
\textsc{Capped-Plain} $(30$--$60)$ and $(20$--$45)$ are the identical prompt
with only the word range changed, targeting the human mean ($54$ words) and
median ($36$ words) respectively. Shortening the budget raises ROUGE-L and
BERTScore for every model without altering anything else, and under the
matched budgets six of the seven models exceed the human agreement baseline
on BERTScore ($0.114$); the exception, LLaVA-1.6-Mistral-7B, is also the one
model that fails to comply with the budget ($117.7$ and $96.6$ words).
Reference-based similarity on this task is therefore driven by length
compatibility as much as by critique content.

\clearpage

\section{Limitation and Future Works}

Our human references are community critiques from \texttt{r/photocritique},
not expert annotations. They are the feedback photographers actually receive,
which is what we want to compare against, but they are often partial, casual,
and uneven in quality, so ``human-like'' in our analysis means resembling this
particular population rather than meeting an expert standard. Relatedly, our
benchmark covers English-language posts, eight open-weight models and a single
closed-weight system; other proprietary models may behave differently, and we
make no claims beyond the models and language we evaluate.

\vspace{0.2cm}
Several analyses rely on LLM judges, which are subject to verbosity and
self-preference biases~\cite{dubois2024length,wang2024large,panickssery2024llm}
that we cannot fully separate from genuine quality. Using four judges rather
than one bounds that risk without removing it: they agree on how similar two
critiques are, and with our annotators, but differ on
which critique is more useful, so we report preferences as the judges'
assessment rather than ground truth. Our central claims rest instead on the
reference-based metrics, the grounding controls, and the behavioral measures
of length, coverage, and diversity. Aspect coverage is itself obtained by
automated tagging, validated in Appendix~\ref{sec:app-tagger} against an
independent labeling but shallow by construction, so the coverage gaps are
upper bounds. Throughout we characterize how model critiques differ from human
ones without isolating the cause; the divergence is consistent with
instruction tuning toward comprehensive, affirmative responses, but our
experiments do not establish that mechanism.

\vspace{0.2cm}
Finally, our results are diagnostic rather than prescriptive. We show that
reference-based similarity is a poor proxy for open-ended critique quality,
and that a single LLM judge is not a reliable substitute, but we do not offer
a metric to replace either. A natural next step is evaluation that targets the
properties these metrics miss: selectivity in choosing what to discuss,
diagnostic specificity, willingness to name
flaws~\cite{torabi2026neuromamballm}, and responsiveness to the question a
poster actually asks. A second is to test our training implication directly,
fine-tuning or selecting models against these references and
metrics and measuring whether
doing so amplifies the stylistic uniformity we observe.

\end{document}